\documentclass[10pt,twocolumn,letterpaper]{article}

\usepackage[pagenumbers]{cvpr} 
\usepackage{bm}
\usepackage{makecell}

\usepackage{multirow}
\definecolor{cvprblue}{rgb}{0.21,0.49,0.74}
\usepackage[pagebackref,breaklinks,colorlinks,allcolors=cvprblue]{hyperref}
\hypersetup{
  pdftitle={Closed-Loop Neural Activation Control in Vision-Language-Action Models},
  pdfauthor={Abhijith Babu and Ramneet Kaur and Nathaniel D. Bastian and Olivera Kotevska and Susmit Jha and Yanzhao Wu and Sumit Kumar Jha and Anirban Roy}
}

\def\paperID{7} 
\def\confName{VisCon}
\def\confYear{2026}

\title{Closed-Loop Neural Activation Control in Vision-Language-Action Models} 

\author{Abhijith Babu\\
Florida International University\\
{\tt\small ababu012@fiu.edu}
\and
Ramneet Kaur\\
SRI International\\
{\tt\small ramneet.kaur@sri.com}
\and
Nathaniel D. Bastian\\
United States Military Academy\\
{\tt\small nathaniel.bastian@westpoint.edu}
\and
Olivera Kotevska\\
Oak Ridge National Laboratory\\
{\tt\small kotevskao@ornl.gov}
\and
Susmit Jha\\
SRI International\\
{\tt\small susmitjha@berkeley.edu}
\and
Yanzhao Wu\\
Florida International University\\
{\tt\small yawu@fiu.edu}
\and
Sumit Kumar Jha\\
University of Florida\\
{\tt\small sumit.jha@ufl.edu}
\and
Anirban Roy\\
SRI International\\
{\tt\small anirban.roy@sri.com}
}

\begin{document}
\maketitle
\begin{abstract}

Vision-Language-Action (VLA) models can be steered at test time by intervening on semantically meaningful internal directions, but existing methods use a fixed steering coefficient, effectively operating in open loop. This is poorly suited to embodied control, where task state and concept error evolve over time, often causing overcorrection, oscillation, and reduced task success, especially for temporal behaviors such as speed and smoothness. We propose {CTRL-STEER}, a closed-loop framework that replaces static intervention strength with adaptive, time-varying control signals. The key idea is to decouple representation from regulation: rather than assuming temporal concepts are directly controlled by individual neurons, we steer along motion-aligned residual directions while a feedback controller adjusts intervention magnitude online. We instantiate this framework with both PID and reinforcement learning based controllers. Experiments with a fine-tuned OpenVLA policy on four LIBERO task suites show that {CTRL-STEER} achieves more stable concept regulation and a better steering-task success trade-off than fixed-coefficient baselines, without modifying or retraining the base model.

\end{abstract}

\section{Introduction}

We propose a framework for controlled steering of vision-language-action (VLA) models~\cite{kim2024openvla, black2024pi_0} for inference-time adaptation. VLA models jointly process visual observations and language instructions to generate low-level control actions for robotic agents, unifying perception, semantic reasoning, and control within an end-to-end trainable architecture. A typical VLA model combines a pretrained vision-language model (VLM) as a perceptual-semantic backbone with an action decoder that generates action sequences conditioned on the task instruction~\cite{zitkovich2023rt}. The VLM encodes scene context and task intent, while the action head autoregressively predicts action tokens. Such models exhibit strong semantic generalization, enabling transfer across object categories and robust interpretation of novel natural-language instructions by leveraging pretrained vision-language representations~\cite{kawaharazuka2025vision, impressive_vlas, black2024pi_0}. However, they remain limited in their ability to generalize across physical and geometric variations, including speed (fast vs. slow), spatial attributes (height, corners, edges), and fine-grained configuration changes~\cite{zhao2025upvla, generalization_problem1, generalization_problem2, generalization_problem3, generalization_problem4, bridgedata, khazatsky2024droid}. These factors require precise grounding in continuous control and system dynamics, where current VLA architectures still struggle.

\begin{figure*}[t]
  \centering
  \includegraphics[width=\textwidth]{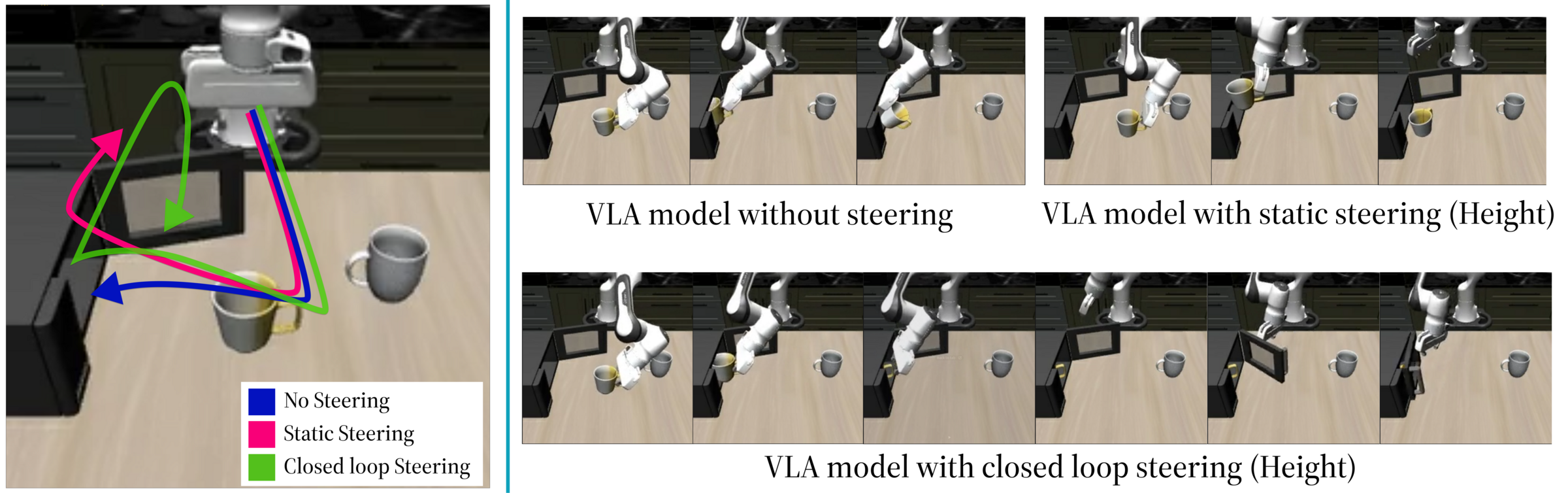}
  \caption{ \textbf{Left}: Example of 
  steering the height concept for the
  task: \textit{Put the yellow and white mug in the microwave and close it.} The goal is to steer the arm to the desired direction, e.g., a larger height.
  \textbf{Right}:
  i. The unsteered VLA model follows a low trajectory and fails to place the mug correctly.
  ii. Static steering of VLA drastically increases the trajectory height and collides with the microwave top.
  iii. CTRL-STEER - our closed-loop RL-based steering dynamically adjusts neuron coefficients, achieving sufficient vertical clearance for successful placement while enabling correct door closure.
  }
  \label{fig:overview}
\end{figure*}

To address this limitation, we propose a controlled steering framework for VLA models that explicitly balances task completion with steering objectives. As illustrated in~\cref{fig:overview} (Left), our method steers the robot arm to move higher while preserving task execution relative to the baseline VLA policy. We formulate this as an inference-time adaptation problem that does not require retraining the underlying VLA model. This design is motivated by the cost and limited scalability of collecting training trajectories that cover diverse physical configurations, especially in novel environments~\cite{huo2025gigabrain, park2025scalable, training1, training2, training3}. Steering VLA models introduces two core challenges: (1) identifying neurons associated with a desired steering instruction, and (2) modulating those neurons to influence predicted actions without compromising task success. The first challenge lies in mechanistic interpretability~\cite{geva2021transformer}, whose goal is to connect semantic concepts to internal neural representations. This is inherently difficult because neurons are often polysemantic, meaning that interpretable features may be distributed across multiple units rather than localized to a single neuron~\cite{sae}. Once such neurons are identified, the second challenge is to regulate their activations so as to produce the desired behavioral change while preserving successful task execution. Existing methods~\cite{haon2025mechanistic, khan2025controlling} typically adopt an open-loop strategy, scaling neuron activations with fixed coefficients to induce target behaviors, but without accounting for the downstream impact on task completion. As a result, improved steering often comes at the expense of degraded task success, exposing the limitations of open-loop intervention when both objectives must be satisfied simultaneously. This is illustrated in \cref{fig:overview} (Right), where steering improves the target behavior but often leads to unsuccessful execution unless task completion is explicitly incorporated. To overcome these challenges, we propose CTRL-STEER, a framework that first identifies neurons associated with continuous-control-relevant concepts (e.g., motion attributes), and then performs closed-loop, task-aware steering of their activations during action generation.

To identify relevant neurons, we build on the mechanistic interpretability approach of~\cite{haon2025mechanistic} for single control features such as up, down, left, right, forward, and backward. Specifically, we project each neuron’s value vector in the feed-forward layer into vocabulary space and derive a semantic embedding that captures its dominant directional meaning. We then select neurons whose embeddings are most similar to a representative {\textit{set of features} associated with the control concept of interest}, yielding a candidate steering set. {This differs from the underlying approach~\cite{haon2025mechanistic}, which performs steering on neurons identified for an individual feature rather than over a set of features relevant to a broader control concept.}

\begin{figure*}[t]
  \centering
  \includegraphics[width=\textwidth]{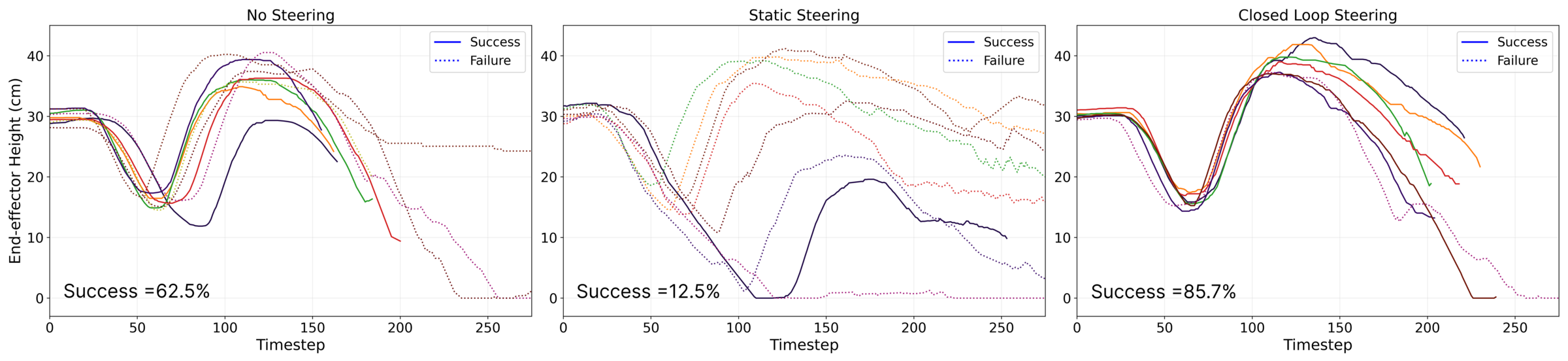}
  \caption{We present the effect of steering height for the task \textit{`pick up the book and place it in the back compartment of the caddy'}. The unsteered model keeps the end-effector lower. Static steering increases the height, but fails to achieve a high task success rate. Closed-loop steering maintains a higher end-effector trajectory while still completing the task.}
  \label{fig:resultanalysis}
\end{figure*}

For closed-loop steering, we employ a proportional-integral-derivative (PID) controller~\cite{astrom1995pid} to regulate neuron activations toward a desired steering objective. Concretely, the PID controller adjusts the scaling factors applied to the identified neurons, thereby shaping the generated action sequence online. Its proportional term corrects instantaneous error, the integral term compensates for accumulated past deviations, and the derivative term responds to the rate of change of the error. While this feedback mechanism improves alignment with the target behavior by incorporating historical error information, it does not explicitly reason about future task outcomes. Consequently, although PID improves steering performance, task success can remain suboptimal. To introduce planning into the closed-loop framework, we further propose a reinforcement learning (RL)-based controller~\cite{rl_control}. Specifically, we train a proximal policy optimization (PPO)~\cite{ppo} policy using a reward that jointly captures steering compliance and task success. Unlike PID, the RL controller learns a nonlinear control policy that adaptively modulates neuron activations over time, enabling coordinated optimization of both steering objectives and overall task performance. As shown in \cref{fig:resultanalysis}, open-loop control suffers from a fundamental steering-success trade-off, whereas closed-loop steering achieves stronger task success while maintaining the desired steering behavior.

We make the following contributions for steering VLA models.

\begin{enumerate}[topsep=0.3pt]
\item \textbf{Closed-Loop Steering.}
We formulate activation steering in VLA models as a closed-loop control problem. Rather than steering individual feature neurons for a continuous control concept using a fixed coefficient, we regulate a set of concept-related features using a PID controller that dynamically adjusts steering coefficients at each timestep.

\item \textbf{RL-based Closed-Loop Steering.}
We introduce an RL-based controller that learns a nonlinear policy for modulating neuron activations over time, jointly optimizing concept steering and task completion.

\item \textbf{Experimental Evaluation.}
Experiments on the LIBERO benchmark show that the closed-loop PID controller maintains task success close to the unsteered baseline (71.37\%), whereas static steering with strong interventions reduces success to as low as 1.8\%. The RL controller further improves success to 73.88\% for height steering and 76.12\% for speed steering while maintaining the desired steering behavior.
\end{enumerate}

\section{Related Work}
\label{sec:relw}

Vision-Language-Action (VLA) models have recently emerged as a powerful paradigm for general-purpose robotic policies by unifying visual perception, language understanding, and action generation within a single model \cite{kim2024openvla, zitkovich2023rt}. Early work showed that pretrained vision-language models (VLMs) can be adapted for robot control by discretizing continuous actions into token sequences, thereby enabling autoregressive action generation \cite{zitkovich2023rt}. Building on this idea, large-scale systems such as OpenVLA \cite{kim2024openvla} and $\pi_0$ \cite{black2024pi_0} combine pretrained VLM backbones with robot interaction data, reflecting a broader shift toward foundation models for robotics. By leveraging large pretrained language backbones, such as LLaMA \cite{llama2}, together with visual encoders, these models enable semantic reasoning to directly shape control behavior.

\textbf{Mechanistic Interpretability.}
Mechanistic interpretability aims to understand neural networks by analyzing their internal representations and identifying semantically meaningful computational components \cite{geva2021transformer, olah2017feature}. In transformer architectures, feedforward layers have been interpreted as key-value memories, where individual neurons correspond to interpretable features in vocabulary space \cite{geva2021transformer, geva2022transformer}. More recently, sparse autoencoder methods have been proposed to decompose activations into more monosemantic directions \cite{bricken2023monosemanticity,sae}. Together, these findings suggest that high-level semantic concepts can often be localized within specific activation subspaces of large models, providing a basis for interpretable intervention.

\textbf{Activation-Level Steering and Representation Engineering.}
Building on these interpretability insights, recent work has explored activation-level steering as a means of modulating model behavior at inference time. In language models, representation engineering methods identify latent directions associated with high-level concepts and steer model outputs by injecting or suppressing those directions \cite{meng2022locating, zou2023representation}. Extending this idea to robotics, \cite{haon2025mechanistic} introduced a method for steering VLA models by identifying semantically aligned feedforward neurons and modulating their activations. Their results demonstrated zero-shot behavioral modulation in both simulation and real-world robots, indicating that control-relevant semantic structure persists within pretrained representations. However, these interventions are typically open-loop: steering coefficients are fixed or manually chosen and remain constant throughout execution. Consequently, they do not account for task-level feedback or the evolving tradeoff between concept steering and task completion.

\textbf{Feedback Control in Robotics.}
In contrast, classical robotics systems rely on closed-loop control to regulate behavior under uncertainty. Controllers such as PID and reinforcement learning (RL)-based policies adapt control signals online using observed errors and task objectives \cite{aastrom2008analysis, aastrom2021feedback}. These feedback mechanisms enable agents to respond to environmental variation, correct deviations during execution, and balance competing objectives more effectively.

Our work connects activation-level steering with classical feedback control by formulating neuron-level concept modulation in VLA models as a closed-loop control problem. Rather than applying fixed steering coefficients, we introduce adaptive controllers that regulate internal activations during inference, enabling a better balance between semantic concept alignment and task success.

\section{Approach}

\begin{figure*}[t]
  \centering
  \includegraphics[width=0.9\textwidth]{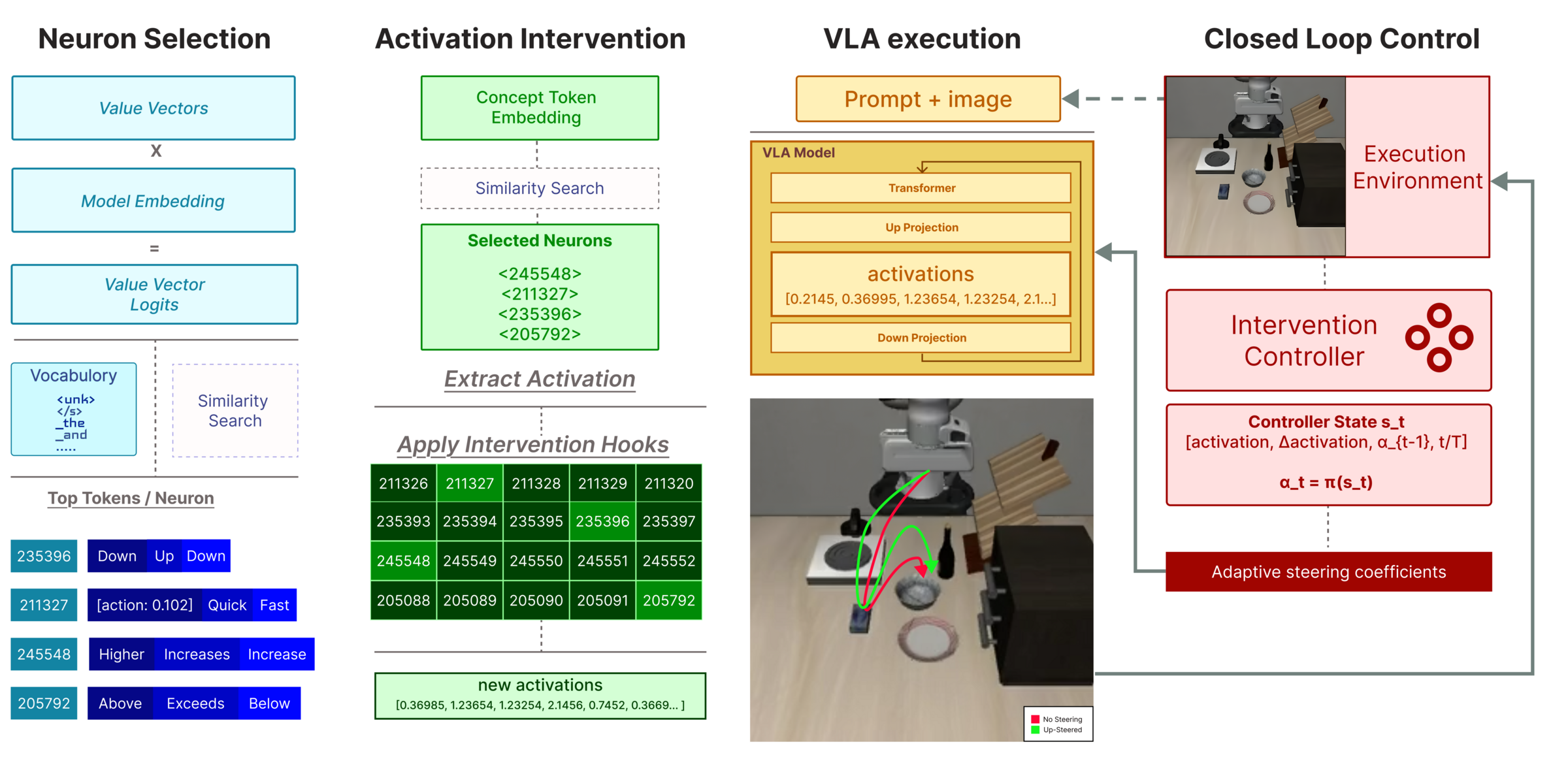}
  \caption{We present CTRL-STEER for controlled steering of VLA models. First, we identify feature-aligned neurons by projecting FFN value vectors into token space and selecting neurons whose semantic embeddings align with representative control features. During inference, these neurons are modulated through activation intervention. A closed-loop controller then adaptively regulates the steering signal using environmental feedback and task rewards, enabling dynamic behavior control while maintaining high task success during VLA execution.}
  \label{fig:framework}
\end{figure*}

We formulate steering in VLA models as a closed-loop control problem, where concept-relevant neuron activations are dynamically regulated to induce desired motion behaviors while preserving task performance. Our method consists of two components. First, we identify neurons associated with control-relevant features using a mechanistic interpretability procedure. Second, we introduce two closed-loop controllers—a PID controller and a reinforcement learning (RL) controller—that adaptively modulate these neurons during inference to balance steering and task execution. An overview of the framework is shown in~\cref{fig:framework}.

\subsection{Mechanistic Interpretability of Temporal Concepts}

Our mechanistic interpretability approach builds on the framework of Haon et al.~\cite{haon2025mechanistic}, which identifies feed-forward network (FFN) neurons associated with a single feature corresponding to a control concept.

\paragraph{FFN decomposition~\cite{geva2021transformer}.}
Consider a transformer block at layer $\ell$, and let $r^{\ell} \in \mathbb{R}^{d}$ denote the residual-stream representation entering the FFN sublayer, where $d$ is the residual dimension. The FFN produces an update $\mathrm{FFN}^{\ell}(r^{\ell}) \in \mathbb{R}^{d}$. Following the standard two-layer FFN structure, this update can be written as
\begin{align}
  \mathrm{FFN}^{\ell}(r^{\ell})
  =
  \sum_{i=1}^{d_m}
  m_i^{\ell} \, v_i^{\ell},
\end{align}
where $d_m$ is the intermediate FFN width, $m_i^{\ell} \in \mathbb{R}$ is the input-dependent activation of neuron $i$ in layer $\ell$, and $v_i^{\ell} \in \mathbb{R}^{d}$ is its input-independent value vector. Each neuron $(\ell,i)$ is therefore associated with a unique value vector $v_i^{\ell}$, which determines the direction of its contribution to the output residual stream. Because these value vectors lie in the same space as the transformer residual stream, they can be projected directly into vocabulary-logit space through the model output head.

\paragraph{Semantic characterization of neurons~\cite{haon2025mechanistic}.}
To assign semantic meaning to individual neurons, we analyze their value vectors in vocabulary space. Let $W_{\mathrm{out}} \in \mathbb{R}^{|\mathcal V| \times d}$ denote the LM head that maps residual representations to logits over the vocabulary $\mathcal V$, where $|\mathcal V|=32{,}000$ for OpenVLA. For each value vector $v_i^{\ell}$, we compute
\begin{align}
z_i^{\ell} = W_{\mathrm{out}} v_i^{\ell} \in \mathbb{R}^{|\mathcal V|},
\end{align}
and convert it to a probability distribution
\begin{align}
p_i^{\ell} = \mathrm{softmax}(z_i^{\ell}).
\end{align}
This distribution reflects which tokens are promoted when the direction $v_i^{\ell}$ is added to the residual stream. We then select the top-$k$ tokens from $p_i^{\ell}$, with $k=20$, and compute their embeddings using the model text encoder. Let $e(w) \in \mathbb{R}^{d_e}$ denote the embedding of token $w$. The semantic embedding of neuron $(\ell,i)$ is defined as the probability-weighted average
\begin{align}
  sem_i^{\ell}
  =
  \sum_{w \in \mathrm{TopK}(p_i^{\ell})}
  p_i^{\ell}(w)\, e(w),
\end{align}
which captures the dominant linguistic meaning associated with $v_i^{\ell}$.

Unlike prior work~\cite{haon2025mechanistic}, which identifies neurons for a single feature, our objective is to construct a steering set aligned with a broader control-relevant concept expressed through multiple related features. We therefore use the semantic embeddings $sem_i^{\ell}$ to retrieve neurons whose value directions consistently align with a representative feature set for the target concept.

\paragraph{Concept-to-neuron matching.}
Given a target control concept $\mathcal C$, we define a set of representative feature tokens
$\mathcal T_{\mathcal C} \subset \mathcal V$
that express the concept. For motion, we use
\{\texttt{up}, \texttt{down}, \texttt{left}, \texttt{right}, \texttt{forward}, \texttt{backward}\}.
For each token $w \in \mathcal T_{\mathcal C}$, we compute its embedding $e(w)$ and perform a $k$-nearest-neighbor (k-NN) search over the set $\{sem_i^{\ell}\}$ across all FFN layers using cosine similarity. For each representative token, the $k$ most similar neurons are selected, and the final candidate set is defined as
\begin{align}
\mathcal S
=
\bigcup_{w \in \mathcal T_{\mathcal C}}
\mathrm{kNN}_k\!\left(e(w)\,;\,\{sem_i^{\ell}\}_{\ell,i}\right).
\end{align}
Although the search spans all layers, we empirically observe that the selected neurons are concentrated mostly in the latter half of the transformer, suggesting stronger semantic alignment in higher layers.

\paragraph{Polysemantic filtering.}
Transformer neurons are often polysemantic, meaning that a single neuron may capture multiple unrelated features~\cite{sae}. We therefore refine $\mathcal S$ by manually inspecting the top-$k$ promoted tokens for each candidate neuron. Neurons whose promoted tokens reflect unrelated or conflicting concepts are removed. After filtering, we retain the 10 neurons with the strongest and most consistent semantic alignment, forming the final intervention set $\mathcal S$.

Control concepts in robotic policies can be broadly divided into \textit{state-based} and \textit{temporal} concepts. State-based concepts correspond to instantaneous properties such as end-effector height, object position, or spatial relation. These can often be represented as directions in the residual stream, which motivates existing approaches based on neuron selection and static steering through value-vector alignment. Temporal concepts, by contrast, such as speed and acceleration, depend on the evolution of behavior across timesteps. Since the VLA policy is invoked independently at each timestep and computes residual updates only from the current observation, these temporal quantities cannot be computed within a single forward pass. As a result, correlation between a value vector and a temporal metric does not imply direct controllability. This motivates our formulation of steering as a closed-loop control problem in which intervention strength is regulated dynamically during execution.

\subsection{Steering as a Closed-Loop Control Problem}

Prior work~\cite{haon2025mechanistic} intervenes on selected neurons by replacing their activation coefficients with a constant scalar $\alpha \in \mathbb{R}$:
\begin{align}
  \tilde m_i^{\ell}
  =
  \begin{cases}
    \alpha & \text{if } (\ell,i) \in \mathcal S, \\
    m_i^{\ell} & \text{otherwise}.
  \end{cases}
\end{align}
This induces a residual shift in the span of
$\{ v_i^{\ell} : (\ell,i) \in \mathcal S \}$,
thereby biasing the action-token distribution toward the target concept in an open-loop manner. While constant activation interventions can produce steering, they often degrade task completion because the steering coefficient does not adapt to the evolving state of the environment. Such static modulation is especially ineffective for temporally varying attributes such as speed or smoothness, where execution history and accumulated deviation matter.

We therefore pose steering as a closed-loop control problem. Let $c^t$ denote the value of a target concept at timestep $t$, and define the steering error as
\begin{align}
    e^t = c^* - c^t,
    \label{eq:error}
\end{align}
where $c^*$ is the desired target value for the concept. The specific choice of $c^*$ depends on the steering objective; for example, for height steering we set $c^*$ to twice the initial height. Instead of using a constant coefficient $\alpha$, we estimate a time-varying steering vector $\bm{\alpha}^t$, with separate steering coefficients for the selected neurons. The closed-loop steering problem is formulated as
\begin{equation}
    \bm{\alpha}^t = \underset{\bm{\alpha} \in \mathcal{R}^k}{\arg \min} \sum_{\tau = t+1}^{T} e^\tau,
\end{equation}
where $\bm{\alpha}^t_i$ denotes the steering coefficient for the $i$-th neuron, and $T$ is the control horizon.

\subsubsection{PID-Based Control.}

We first introduce a proportional-integral-derivative (PID) controller to compute the time-varying steering signal $\bm{\alpha}_{PID}^t$ from the steering error in \cref{eq:error}. The controller is defined as
\begin{equation}
    \bm{\alpha}^t_{PID} = K_P \cdot e^t + K_I\sum_{\tau=0}^{t}e^{\tau} + K_D(e^t - e^{t-1}),
    \label{equ:PID}
\end{equation}
where the three terms correspond to the proportional, integral, and derivative components, scaled by coefficients $K_P$, $K_I$, and $K_D$, respectively. The PID controller is independent of the neurons and hence outputs a scalar signal which is applied to all the neurons.

\textbf{Proportional term.}
The proportional term adjusts neuron activations according to the instantaneous concept error. It scales the residual shift along the steering directions in proportion to the current deviation from the desired behavior, enabling rapid correction when the trajectory departs from the target concept.

\textbf{Integral term.}
The integral term accumulates past errors to compensate for persistent bias. Because the base VLA model is optimized for its original training objective, the steering objective may conflict with its learned dynamics, resulting in steady-state deviations. The proportional term alone cannot eliminate such bias. The integral term progressively strengthens the intervention until the accumulated error is reduced.

\textbf{Derivative term.}
The derivative term responds to the rate of change of the steering error. When deviations increase rapidly, it amplifies the control signal to accelerate correction; when the error decreases quickly, it attenuates the intervention to avoid overshoot. Since steering primarily affects motion-related behavior, abrupt changes in neuron scaling can induce oscillatory trajectories. The derivative term helps suppress such oscillations and improves stability.

Together, these components provide stable and responsive closed-loop regulation of the selected steering neurons. To avoid excessive residual shifts, we constrain $\bm{\alpha}^t_{PID} \in [0,20]$, based on empirical observations that stronger interventions degrade task performance. The gains $K_P$, $K_I$, and $K_D$ are chosen conservatively to avoid over-amplification, especially in the presence of noisy error signals that can disproportionately affect the derivative term.

\subsubsection{RL-Based Control}

Although the PID controller adapts neuron activations using past error, it remains fundamentally reactive and does not explicitly optimize long-horizon task success. As a result, it can improve steering behavior without fully preserving task completion. To address this limitation, we introduce a reinforcement learning (RL)-based controller that performs long-term planning and jointly optimizes steering compliance and task success. Rather than modulating neurons for a single isolated feature, the RL controller operates over a set of neurons associated with a broader collection of motion-related concepts. This allows it to model nonlinear dependencies across concept-aligned neurons and coordinate their activations more effectively.

We formulate steering as a reinforcement learning problem in which a policy predicts neuron activations that induce the desired behavior. At timestep $t$, the controller state is defined as
\begin{align}
    s_t = [a_t, \Delta a_t, \alpha^{t-1}, \dfrac{t}{T}],
\end{align}
where $a_t \in \mathbb{R}^k$ denotes the activations of the $k$ selected neurons, $\Delta a_t = a_t - a_{t-1}$ captures the immediate change in neuron activations, $\alpha^{t-1} \in \mathbb{R}^k$ is the steering signal applied at the previous timestep, and $t/T$ is the normalized episode progress.

The policy $\pi_{\theta}$ outputs
\begin{align}
    \bm{\alpha}^t_{RL} = \pi_{\theta}(s_t),
\end{align}
where $\bm{\alpha}^t_{RL}$ is a $k$-dimensional vector assigning steering values to the selected neurons. These values directly modulate FFN activations and thereby steer the VLA toward the desired concept. The reward function is designed to balance steering quality and task success:
\begin{align}
    r_t = r_{steer}(t) + \lambda \cdot r_{task},
    \label{eq:rlreward}
\end{align}
where $r_{steer}$ measures instantaneous steering metric, such as height or instantaneous velocity, and $r_{task}$ is the task-completion reward. The coefficient $\lambda$ determines the trade-off between the two objectives.

We train a separate RL policy for each task. For a given task, the policy $\pi_{\theta}$ is optimized using rollouts collected from the corresponding environment. We initialize the policy using the steering signal $\bm{\alpha}^t_{PID}$ generated by the PID controller, and then further refine it through environment interaction using rewards that jointly encode concept alignment and task completion. This task-specific optimization enables the controller to adapt neuron activations to the dynamics of each task while balancing steering objectives with successful execution.

\subsubsection{Implementation Details.}

We use OpenVLA~\cite{kim2024openvla} that is trained on prismatic VLM~\cite{karamcheti2024prismatic}, that uses LLaMA-2~\cite{llama2} as the language backbone. For neuron identification, we consider all FFN layers in the LLaMA-2 decoder and select 10 neurons associated with the motion-related concept. In the concept-to-neuron matching step, we set $k=5$ for the k-NN search. For height steering, the target value is set to $c^{*}=2h_{0}$, where $h_{0}$ is the initial end-effector height. For speed steering, we set $c^{*}=30$ cm/s. For the PID controller, we use gains $K_P=4.0$, $K_I=0.5$, and $K_D=1.0$ with a control horizon of 20 steps. For the RL-based controller, we use PPO for policy optimization. Each trajectory has a horizon of $T=920$ steps, consisting of 20 environment warm-up steps followed by 900 execution steps for all tasks in the LIBERO benchmark.
\section{Experiments}

\textbf{Benchmark.} 
We evaluate our method on the LIBERO benchmark and BridgeData V2\cite{bridgedata}. We consider all four task suites from the LIBERO \cite{liu2023libero} benchmark: LIBERO-Goal, LIBERO-Object, LIBERO-Spatial, and LIBERO-Long. LIBERO-Goal has 10 tasks that focus on direct goal manipulation, LIBERO-Object has 10 tasks focusing on object-centric variations, and LIBERO-Spatial has 10 tasks considering spatial relations. For LIBERO-Long, we use the libero-10 subset that consists of long-horizon sequential tasks. From BridgeData V2, we evaluated on pick-and-place tasks on SimplerEnv \cite{li2024evaluating} These tasks encompass compositional complexity, reasoning, and motion diversity, making it a suitable benchmark for evaluating state-based and temporal steering of VLAs.\\

\begin{table}[t]
\centering
\caption{Steering height concept in X-VLA~\cite{zheng2025x} model on the LIBERO Goal task suite. AAT measures the area above the threshold height, showcasing the steerability, and SR shows the task success rate.}
\resizebox{\columnwidth}{!}{%
\label{tab:xvla_height_libero_goal}
\begin{tabular}{lcccc}
\toprule
Method & Height & 95th Percentile & AAT & SR(\%) \\
\midrule
Unsteered & $1.041$ & $1.140$ & $161.58$ & $59\%$ \\
Static 20 & $1.020$ & $1.073$ & $626.74$ & $25\%$ \\
PID       & $1.046$ & $1.087$ & $183.35$ & $60\%$ \\
RL        & $1.037$ & $1.123$ & $400.33$ & $60\%$ \\
\hline
\bottomrule
\end{tabular}
}
\end{table}

\textbf{Baseline VLA model.}
We use OpenVLA \cite{kim2024openvla} as the base VLA policy. OpenVLA is a 7B-parameter model with llama2 as its backbone. For each task suite, OpenVLA is fine-tuned independently to improve its task completion capabilities, without making architectural modifications. We additionally evaluate on X-VLA~\cite{zheng2025x} to showcase the transferability of our method on other VLA architectures.\\

\begin{table}[t]
\centering
\caption{ \small OpenVLA performance in terms of height and speed steering and success rate (SR) on BridgeData V2.}
\resizebox{\columnwidth}{!}{%

\label{tab:openvla_height_bridge}
\begin{tabular}{lcccc}
\toprule
Method & AAT (height) & SR (Height) & Avg speed & SR (Speed)\\
\midrule
Unsteered   & $39.15$ & $40\%$ & 16.93 & 40\% \\ 
Static (20) & $47.96$ & $12.5\%$ & 25.18 & 14.9\% \\
PID         & $47.28$ & $46\%$ & 18.90 & 45.8 \% \\
RL          & $48.13$ & $47.6\%$ & 21.38 & 48.9\% \\
\hline
\bottomrule
\end{tabular}
}
\end{table}

\textbf{Steerable Concepts.}
We consider two concepts for steering: (1) \textit{Height} is chosen as a state-based concept determined by the instantaneous height of the end-effector. Height represents a validation case for spatial concepts, where we expect our framework to perform well, in addition to temporal concepts. (2)\textit{ Speed} of the end-effector across time steps is chosen as the temporal control concept.\\
\textbf{Metrics.} 
We evaluate steering using concept-specific metrics. For height, we report average end-effector height \cite{haon2025mechanistic}, $95^{th}$ percentile height, and the Area Above Threshold (AAT). For all metrics, height is measured in meters. For speed, we report the average end-effector speed\cite{haon2025mechanistic} and the Speed Above Threshold (SAT), with all speed values measured in cm/s.
\begin{table*}[t]
\caption{Comparative analysis of spatial steering for end-effector height across LIBERO task suites. We evaluate the unsteered OpenVLA model against static steering ($C \in \{5, 10, 20\}$) and CTRL-STEER: our closed-loop steering approach with PID and RL controllers. Height (m) measures the distance of the end-effector from ground, AAT represents the Area Above Threshold, and SR (\%) denotes the task success rate. CTRL-STEER demonstrates a trade-off between concept adherence and success rate.}
\centering
\small
\scalebox{0.86}{
\begin{tabular}{llcccc}
\toprule
Task Suite & Method & Height (m) & $95^{th}$ percentile & AAT  & SR (\%) \\
\toprule
\multirow{6}{*}{LIBERO LONG} & OpenVLA\cite{kim2024openvla} & 0.832 $\pm$  0.249 & 0.940 $\pm$  0.259 & 181.655 $\pm$  107.034 & \textbf{58.00} \\
& C=5\cite{haon2025mechanistic} & 0.816 $\pm$  0.252 & 0.923 $\pm$  0.260 & 169.030 $\pm$  113.953 & 52.50 \\
& C=10\cite{haon2025mechanistic} & 0.817 $\pm$  0.250 & 0.924 $\pm$  0.258 & 192.421 $\pm$  128.940 & 51.00 \\
& C=20\cite{haon2025mechanistic} & 0.723 $\pm$  0.213 & 0.841 $\pm$  0.221 & 201.075 $\pm$  136.052 & 10.00 \\
& CTRL-STEER(PID) & 0.828 $\pm$  0.249 & 0.937 $\pm$  0.262 & 184.498 $\pm$  130.825 & 54.00 \\
& CTRL-STEER(RL) & 0.804 $\pm$  0.252 & 0.914 $\pm$  0.260 & 181.858 $\pm$  122.344 & 57.00 \\
\midrule
\multirow{6}{*}{LIBERO GOAL} & OpenVLA\cite{kim2024openvla} & 1.056 $\pm$  0.053 & 1.194 $\pm$  0.052 & 107.764 $\pm$  89.061 & 77.50 \\
& C=5\cite{haon2025mechanistic} & 1.060 $\pm$  0.052 & 1.196 $\pm$  0.046 & 104.280 $\pm$  74.879 & 79.00 \\
& C=10\cite{haon2025mechanistic} & 1.057 $\pm$  0.051 & 1.197 $\pm$  0.046 & 98.358 $\pm$  71.684 & 77.00 \\
& C=20\cite{haon2025mechanistic} & 1.041 $\pm$  0.053 & 1.187 $\pm$  0.059 & 95.922 $\pm$  91.429 & 41.00 \\
& CTRL-STEER(PID) & 1.058 $\pm$  0.049 & 1.194 $\pm$  0.045 & 108.788 $\pm$  98.250 & 79.00 \\
& CTRL-STEER(RL) & 1.062 $\pm$  0.052 & 1.200 $\pm$  0.046 & 99.779 $\pm$  61.452 & \textbf{82.00} \\
\midrule
\multirow{6}{*}{LIBERO OBJECT} & OpenVLA\cite{kim2024openvla} & 0.212 $\pm$  0.032 & 0.293 $\pm$  0.027 & 4.440 $\pm$  5.107 & 72.00 \\
& C=5\cite{haon2025mechanistic} & 0.217 $\pm$  0.026 & 0.293 $\pm$  0.025 & 4.769 $\pm$  5.364 & 72.50 \\
& C=10\cite{haon2025mechanistic} & 0.211 $\pm$  0.029 & 0.293 $\pm$  0.028 & 4.661 $\pm$  6.693 & 73.00 \\
& C=20\cite{haon2025mechanistic} & 0.203 $\pm$  0.034 & 0.300 $\pm$  0.031 & 2.596 $\pm$  2.390 & 33.50 \\
& CTRL-STEER(PID) & 0.210 $\pm$  0.033 & 0.295 $\pm$  0.029 & 4.496 $\pm$  6.324 & 76.00 \\
& CTRL-STEER(RL) & 0.211 $\pm$  0.030 & 0.292 $\pm$  0.026 & 4.671 $\pm$  6.225 & \textbf{77.00} \\
\midrule
\multirow{6}{*}{LIBERO SPATIAL} & OpenVLA\cite{kim2024openvla} & 1.068 $\pm$  0.051 & 1.192 $\pm$  0.041 & 90.940 $\pm$  27.236 & 78.00 \\
& C=5\cite{haon2025mechanistic} & 1.071 $\pm$  0.051 & 1.189 $\pm$  0.040 & 87.748 $\pm$  21.590 & 79.00 \\
& C=10\cite{haon2025mechanistic} & 1.073 $\pm$  0.049 & 1.189 $\pm$  0.040 & 110.937 $\pm$  110.628 & \textbf{84.00} \\
& C=20\cite{haon2025mechanistic} & 1.058 $\pm$  0.046 & 1.186 $\pm$  0.035 & 127.337 $\pm$  90.296 & 25.00 \\
& CTRL-STEER(PID) & 1.068 $\pm$  0.047 & 1.187 $\pm$  0.037 & 92.928 $\pm$  58.663 & 75.00 \\
& CTRL-STEER(RL) & 1.072 $\pm$  0.052 & 1.191 $\pm$  0.041 & 97.970 $\pm$  70.173 & 79.50 \\
\midrule
\bottomrule
\end{tabular}
}
\label{tab:height_results}
\end{table*}

\begin{table*}[t]
\centering
\caption{Comparative analysis of temporal steering for end-effector speed across LIBERO task suites. We evaluate the performance of the unsteered OpenVLA baseline against static steering ($C \in \{5, 10, 20\}$) and CTRL-STEER: our closed-loop steering framework utilizing PID and RL controllers. Speed (cm/s) denotes average speed of the end effector, SAT represents the Speed Above Threshold metric to isolate active motion, and SR (\%) denotes the task success rate. CTRL-STEER provides stable speed regulation while significantly mitigating the success rate degradation observed in high-coefficient static interventions.}
\small
\scalebox{0.9}{
\begin{tabular}{llccc}
\toprule
TASK SUITE & METHOD & Speed (cm/s) & SAT (cm/s) & SR (\%) \\
\toprule
\multirow{6}{*}{LIBERO LONG} & OpenVLA\cite{kim2024openvla} & 11.255 $\pm$ 2.701 & 2.688 $\pm$ 1.303 & 58.00 \\
 & C=5\cite{haon2025mechanistic} & 11.297 $\pm$ 2.673 & 2.677 $\pm$ 1.452 & 53.50 \\
 & C=10\cite{haon2025mechanistic} & 11.111 $\pm$ 2.911 & 2.773 $\pm$ 1.317 & 44.00 \\
 & C=20\cite{haon2025mechanistic} & 12.700 $\pm$ 0.320 & 4.270 $\pm$ 0.660 & 1.50 \\
 & CTRL-STEER(PID)  & 10.877 $\pm$ 3.060 & 2.603 $\pm$ 1.340 & 59.50 \\
 & CTRL-STEER(RL)  & 11.037 $\pm$ 2.843 & 2.535 $\pm$ 1.210 & \textbf{66.50 }\\
\midrule
\multirow{6}{*}{LIBERO GOAL} & OpenVLA\cite{kim2024openvla} & 14.022 $\pm$ 4.553 & 3.651 $\pm$ 1.235 & 77.50 \\
 & C=5\cite{haon2025mechanistic} & 14.147 $\pm$ 4.114 & 3.579 $\pm$ 1.418 & 80.50 \\
 & C=10\cite{haon2025mechanistic} & 13.229 $\pm$ 4.600 & 3.587 $\pm$ 1.279 & 59.00 \\
 & C=20\cite{haon2025mechanistic} & 14.516 $\pm$ 2.998 & 3.856 $\pm$ 1.304 & 2.50 \\
 & CTRL-STEER(PID)  & 13.862 $\pm$ 4.634 & 3.624 $\pm$ 1.216 & 76.00 \\
 & CTRL-STEER(RL)  & 14.040 $\pm$ 4.396 & 3.488 $\pm$ 1.388 & \textbf{83.00 }\\
\midrule
\multirow{6}{*}{LIBERO OBJECT} & OpenVLA\cite{kim2024openvla} & 14.156 $\pm$ 3.132 & 2.126 $\pm$ 0.380 & 72.00 \\
 & C=5\cite{haon2025mechanistic} & 14.712 $\pm$ 2.819 & 2.250 $\pm$ 0.412 & 76.00 \\
 & C=10\cite{haon2025mechanistic} & 15.180 $\pm$ 2.502 & 2.301 $\pm$ 0.381 & 63.00 \\
 & C=20\cite{haon2025mechanistic} & 10.160 $\pm$ 3.652 & 2.468 $\pm$ 0.139 & 2.50 \\
 & CTRL-STEER(PID)  & 14.665 $\pm$ 2.698 & 2.189 $\pm$ 0.443 & \textbf{76.50 }\\
 & CTRL-STEER(RL)  & 14.111 $\pm$ 3.149 & 2.171 $\pm$ 0.466 & \textbf{76.50 }\\
\midrule
\multirow{6}{*}{LIBERO SPATIAL} & OpenVLA\cite{kim2024openvla} & 17.519 $\pm$ 2.695 & 3.569 $\pm$ 0.772 & 78.00 \\
 & C=5\cite{haon2025mechanistic} & 17.461 $\pm$ 3.022 & 3.603 $\pm$ 0.785 & 76.50 \\
 & C=10\cite{haon2025mechanistic} & 17.042 $\pm$ 2.979 & 3.338 $\pm$ 0.834 & 65.50 \\
 & C=20\cite{haon2025mechanistic} & 11.855 $\pm$ 0.895 & 6.150 $\pm$ 3.840 & 1.00 \\
 & CTRL-STEER(PID)  & 17.636 $\pm$ 2.631 & 3.579 $\pm$ 0.767 & 78.00 \\
 & CTRL-STEER(RL) & 17.642 $\pm$ 2.845 & 3.606 $\pm$ 0.820 & \textbf{78.50 }\\
\midrule
\bottomrule
\end{tabular}
}

\label{tab:speed_results}
\end{table*}

\subsection{Results.}

\textbf{Height steering.}
Let $h^t$ denote the end-effector height at timestep $t$ in the LIBERO simulator coordinate system. Following \cref{eq:error}, we set $c^t=h^t$ and $e^t=c^*-h^t$. For the PID controller, $c^*$ is set to twice the initial end-effector height, $2h_0$, ensuring a meaningful upward shift while remaining within the robot workspace. In unsteered trajectories, the end-effector never exceeded this bound across all task suites. We verified that varying the multiplier within $[1.8h_0,2.5h_0]$ does not qualitatively affect the steering–success trade-off.

For the RL controller, the instantaneous height $h^t$ is used as the reward at each timestep. The overall reward follows \cref{eq:rlreward}, combining steering reward with the binary task success signal while exploring trade-off coefficients $\lambda \in \{100, 200, 500, 1000\}$. This allows the policy to learn state-dependent interventions that increase trajectory height while focusing on completing the task.

We evaluated steering using mean end-effector height, but found that it is sensitive to low-height phases occurring during grasping or manipulation near the table. Prior work~\cite{haon2025mechanistic} instead uses maximum height, which is sensitive to initial configurations where the end-effector may already start near its highest position. To better capture sustained elevation, we therefore report two complementary metrics: the $95^{th}$ percentile height and the area above threshold (AAT). The $95^{th}$ percentile reflects the highest operational region of the trajectory while mitigating initial-position effects. AAT integrates the end-effector height above a predefined threshold across the trajectory, capturing how long the arm remains elevated during execution. The threshold $h_{th}$ is defined as the midpoint between the minimum and maximum reachable end-effector heights, providing a task-independent measure of consistent elevation. \cref{tab:height_results} compares our performance with the baseline OpenVLA model and the state-of-the-art steering approach~\cite{haon2025mechanistic} for VLA models with different static activation coefficients ($C=5, 10, 20$). We also report task-wise results in the supplementary material.

\textbf{Speed Steering.}
Here, the aim is to increase the end-effector's motion speed during the task execution. Let $s^t$ denote the instantaneous end-effector speed at timestep $t$, representing the concept of interest: $c^t=s^t$. For the PID controller, we set the target speed to $c^*=30$ cm/s, which exceeds the maximum speed observed in unsteered trajectories while remaining within the robot’s feasible operating range. For the RL controller, the instantaneous speed $s^t$ is used as the concept reward.

Mean end-effector speed does not reliably capture steering effects because manipulation phases in the task execution (e.g., grasping) involve extended stationary periods that reduce the average speed. We, therefore, introduce the speed-above-threshold (SAT) metric with SAT defined as $\mathrm{mean}(s^t - s_{thr} \mid s^t > s_{thr})$,
where $s_{thr}$ is the speed threshold. This metric focuses on active motion segments of the trajectory while filtering out slow manipulation phases. We set $s_{thr}=20$ cm/s, which corresponds to the highest speeds observed during stationary manipulation actions such as grasping. \cref{tab:speed_results} compares our performance with the baselines. We also report task-wise results in the supplementary material. 

To evaluate whether CTRL-STEER transfers beyond OpenVLA, we additionally apply the same steering framework to X-VLA~\cite{zheng2025x} on the LIBERO-Goal task suite. As shown in \cref{tab:xvla_height_libero_goal}, CTRL-STEER preserves task success while improving height-related steering metrics compared to the unsteered X-VLA baseline. Static steering achieves a much larger AAT, but substantially reduces task success, showing the same steering--success trade-off observed for OpenVLA.

\textbf{Analysis of the Results.}
The results in \cref{tab:height_results} and \cref{tab:speed_results} reveal a fundamental limitation of static activation steering: increasing the steering strength to enforce a target concept often leads to a degradation in task success rate. Across the tasks, static steering succeeds when larger intervention strengths are applied. However, this improvement comes at the cost of severe reductions in task success. For example, when using a strong static intervention ($C=20$), the \textbf{average task success rate} drops from \textbf{71.37\%} for the unsteered model to \textbf{27.37\%} in height steering and further to \textbf{1.8\%} in the speed steering setting. These results highlight the limitations of the open-loop steering. In contrast, our closed-loop framework maintains task performance close to the unsteered baseline while still enabling effective concept steering. The PID-based controller achieves average success rate of\textbf{ 71\% }for height steering and \textbf{72.5\%} for speed steering. Furthermore, the RL-based controller improves performance, achieving \textbf{73.88\%} as the average success rate for height steering and \textbf{76.12\% } for speed steering, surpassing the unsteered policy.

\textbf{Trade-off between steering and task success.}
A central challenge in activation steering is balancing concept enforcement with task completion. Overly aggressive steering of a concept can interfere with successful execution. Our controlled steering framework addresses this by dynamically adjusting the steering coefficient, reducing intervention strength when strict concept adherence conflicts with task success.

This effect is evident in height steering tasks for the LIBERO-Goal and LIBERO-Long suites for the RL-based controller in CTRL-STEER, where AAT is lower than the unsteered OpenVLA model but the success rate is maintained. For instance, as shown in Fig.~\ref{fig:overview}, when placing a cup inside a microwave, static steering causes the robot to lift the cup excessively, leading to collisions and object drops. CTRL-STEER mitigates this by decreasing the steering strength during insertion, allowing successful placement. Although the achieved height is slightly reduced, task completion is preserved.

A similar trend appears in speed steering and we provide qualitative results in the supplementary material. In cluttered scenes, excessive speed destabilizes object interactions and reduces success rates. Controlled steering attenuates the intervention when high speed becomes detrimental, whereas static steering with large values (e.g., $C=20$) maintains high speeds but significantly degrades performance. These results demonstrate that effective steering requires adaptive regulation of intervention strength rather than constant perturbation, enabling a better trade-off between concept modulation and task success.

\textbf{Reinforcement Learning without PID initialization.}
We analyze the impact of PID-based initialization on RL training. In our default setting, the RL policy is warm-started using steering coefficients from the PID controller before further optimization. When trained from random initialization instead, task success drops from \textbf{75\%} to \textbf{65\%}, and steering quality (SAT) decreases from \textbf{2.48} to \textbf{2.28}, underperforming both the unsteered baseline and other steering methods. These results suggest that PID provides a stable and meaningful initialization, guiding the RL policy toward effective steering dynamics. Without this warm start, the policy struggles to jointly optimize task completion and concept regulation, resulting in longer training time and degraded performance. We present detailed results in the supplementary material.

\section{Conclusion}

We introduce a closed-loop activation steering framework for Vision-Language-Action models that enables controllable steering while preserving task performance. Unlike prior static interventions that degrade execution, we formulate steering as a feedback control problem and dynamically regulate neuron activations using PID and RL-based controllers. This adaptive mechanism improves the trade-off between concept enforcement and task success across LIBERO benchmarks, with the RL controller maintaining performance above the baseline while achieving effective steering. Our approach has limitations: the RL controller is trained per task, neuron identification still involves partial manual filtering, and evaluation is limited to a small set of steering concepts. Future work includes developing task-agnostic controllers, automating neuron disentanglement, and extending to multi-concept control. Overall, our results highlight the potential of combining feedback control with activation-level interventions for improving the reliability and controllability of embodied models.

\section*{Acknowledgment} The authors acknowledge partial support from NSF (IIS-2331908 and OAC-2530965), DARPA (HR00112420004, HR00112490420, and HR00112490424), DoE (DE-SC0024576), and NAIRR Pilot (NAIRR250261).

{
    \small
    \bibliographystyle{ieeenat_fullname}
    \bibliography{main}
}

\clearpage
\appendix

\section{RL Training without PID Initialization}
\label{sec:pid_ablation}

\begin{table*}[!th]
\centering
\caption{Performance of RL controllers trained without PID  initialization for speed steering across all four LIBERO task suites. Compared to the RL+PID controllers used in the main paper, training RL without PID initialization results in lower task success rates (SR), demonstrating the importance of PID initialization for starting with a stable RL training.}
\small
\scalebox{0.9}{
\begin{tabular}{llccc}
\toprule
TASK SUITE & METHOD & Speed (cm/s) & SAT (cm/s) & SR (\%) \\
\toprule
\multirow{2}{*}{LIBERO LONG} & RL + PID & 11.037 $\pm$ 2.843 & 2.535 $\pm$ 1.210 & \textbf{66.50} \\
 & RL (no PID) & 11.267 $\pm$ 2.902 & 2.669 $\pm$ 1.292 & 57.00 \\

\midrule
\multirow{2}{*}{LIBERO GOAL} & RL + PID & 14.040 $\pm$ 4.396 & 3.488 $\pm$ 1.388 & \textbf{83.00} \\
 & RL (no PID) & 14.127 $\pm$ 4.372 & 3.657 $\pm$ 1.306 & 78.00 \\

\midrule
\multirow{2}{*}{LIBERO OBJECT} & RL + PID & 14.111 $\pm$ 3.149 & 2.171 $\pm$ 0.466 & \textbf{76.50} \\
 & RL (no PID) & 14.080 $\pm$ 2.927 & 2.124 $\pm$ 0.434 & 72.00 \\

\midrule
\multirow{2}{*}{LIBERO SPATIAL} & RL + PID & 17.642 $\pm$ 2.845 & 3.606 $\pm$ 0.820 & \textbf{78.50} \\
 & RL (no PID) & 17.600 $\pm$ 2.910 & 3.606 $\pm$ 0.759 & 77.00 \\

\midrule
\bottomrule
\end{tabular}
}

\label{tab:ablation_results}
\end{table*}
In this section, we provide detailed results across all LIBERO task suites to support our claim in the main paper on ``degraded task success and steering performance for RL training initialized without PID data''.

In our default training setup, the RL controller is initialized using steering coefficients generated by the PID controller, which provides a stable starting point for learning. To evaluate the importance of this initialization, we also trained RL policies from random initialization, without using PID-generated trajectories.

\begin{figure*}[t]
  \centering
  \includegraphics[width=\textwidth]{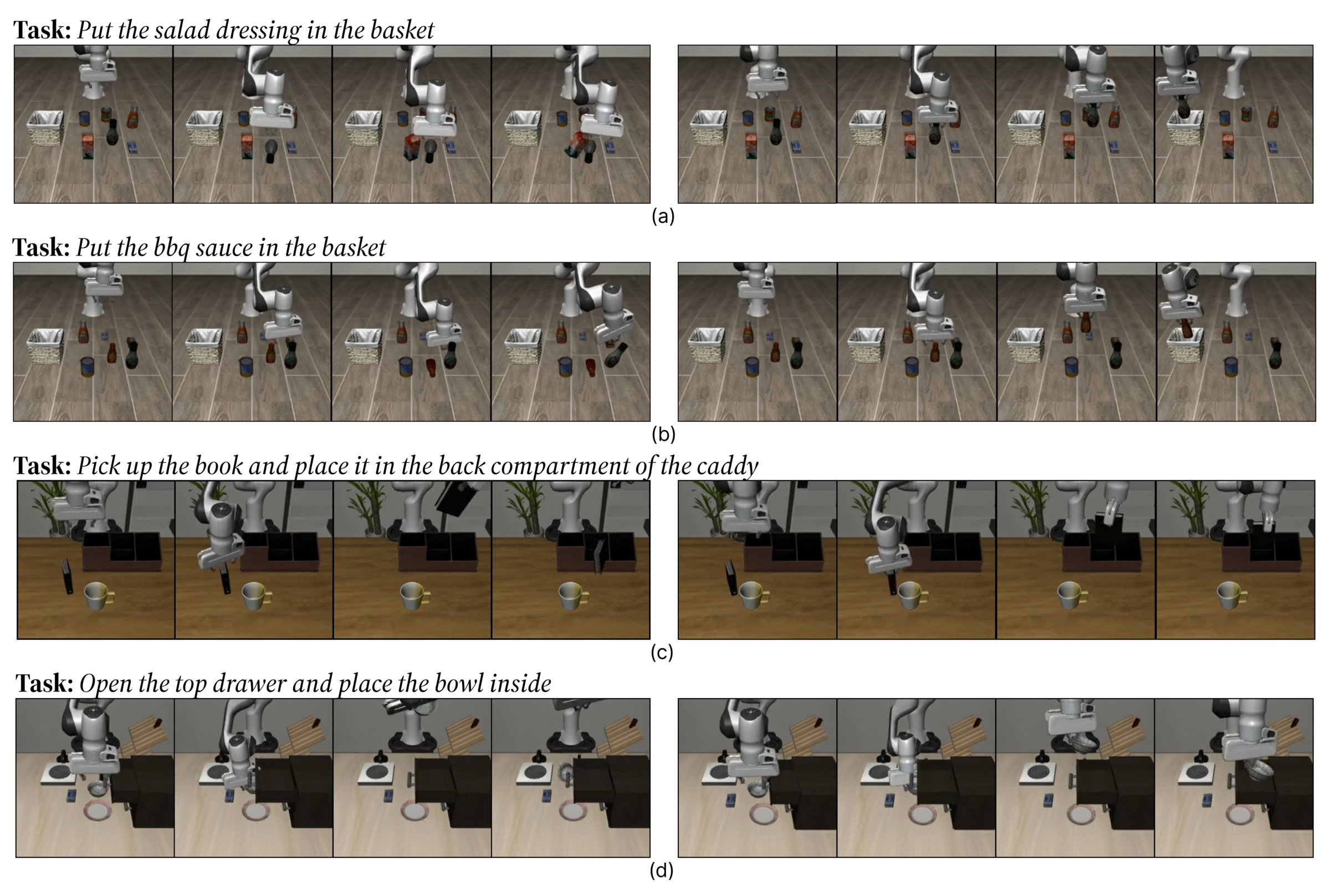}
  \caption{Qualitative examples on steering-success tradeoff that lead to failed tasks in static steering (left)~\cite{haon2025mechanistic}, while achieving success with CTRL-STEER (ours) (right). In (a) and (b), the model was steered for speed, i.e., to move faster. Static steering led to abrupt motion and knocked over the nearby objects. In (c) and (d), the model was steered to go higher. Static steering led to abrupt motion in the upward direction thereby dropping the object. CTRL-STEER adjusts the steering strength and completes the task successfully on these tasks.
}
  \label{fig:tradeoff}
\end{figure*}

\cref{tab:ablation_results} reports the detailed results for speed steering on all four LIBERO task suites: Long, Goal, Object, and Spatial. Across all suites, removing PID initialization consistently leads to lower task success rates while providing no improvement in steering metrics. These results support the conclusion in the main paper that PID initialization stabilizes RL training and improves the trade-off between concept steering and task completion.

\section{Steering Success Tradeoff}
\label{sec:tradeoff}

\begin{figure*}[t]
  \centering
  \includegraphics[width=\textwidth]{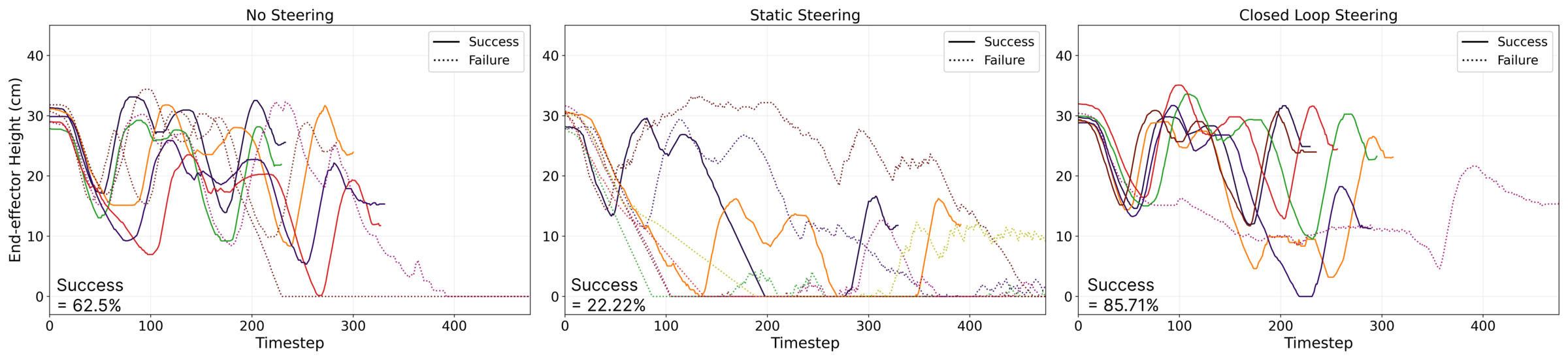}
  \caption{ Effect of steering on the task \textit{`put both the alphabet soup and the cream cheese box in the basket'}. The unsteered model fails in some episodes. Static steering often prevents the robot from successfully grasping the first object, resulting in lower task success and reduced end-effector height. In contrast, closed-loop steering by CTRL-STEER enables successful task completion while maintaining a higher end-effector trajectory whenever possible.
}
  \label{fig:tradeoffplot}
\end{figure*}

In this section, we illustrate qualitative examples of the steering–success trade-off. In \cref{fig:tradeoff}, Static steering with large intervention strengths (left) often produces abrupt motions that destabilize the task execution. In contrast, CTRL-STEER (right) dynamically adjusts the steering coefficient during execution, allowing the robot to maintain the desired concept behavior while preserving task success. \cref{fig:tradeoffplot} shows the effect of steering on the end-effector height where steering by height led to poor task completion. 

\section{Time Complexity Analysis}
\label{sec:timeanalysis}

\cref{tab:complexity_main} shows the average time complexity and peak GPU usage for different approaches. CTRL-STEER imposes low overhead on both time complexity and GPU usage, making it a suitable approach over retraining or finetuning the model. \cref{tab:complexity_rl} to \cref{tab:complexity_naive} shows the detailed computational costs for each approach.

\begin{figure*}[t]
  \centering
  \includegraphics[width=\textwidth]{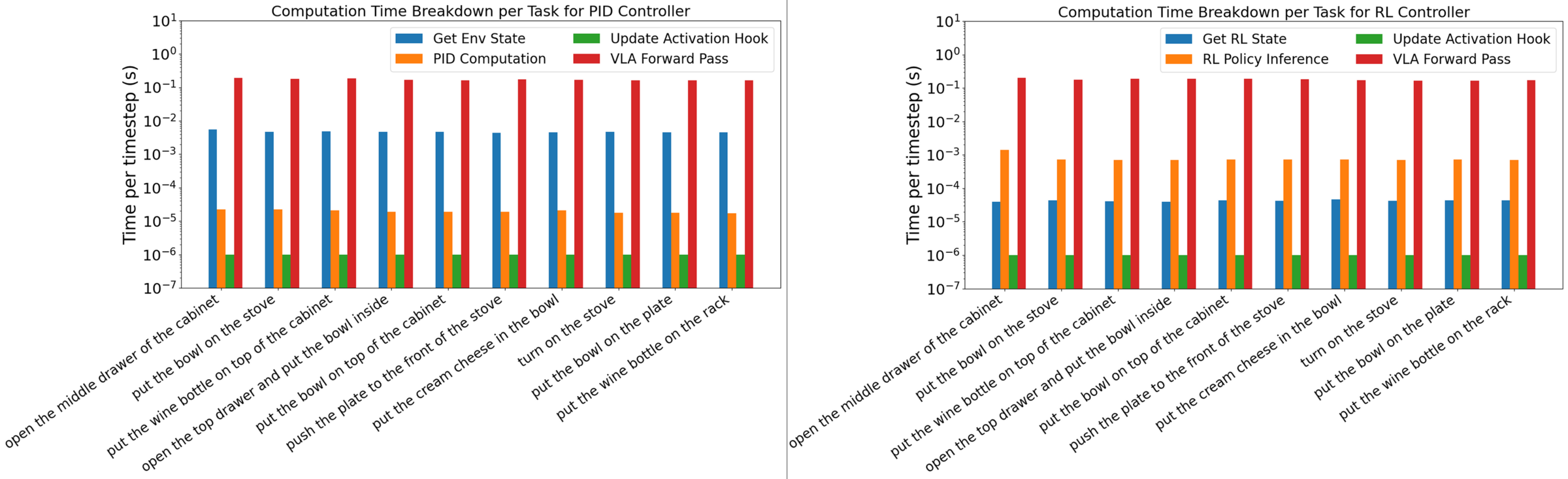}
  \caption{ Breakdown of the per-timestep computational cost for CTRL-STEER with only PID (left) and RL+PID (right). The VLA forward pass dominates the runtime in both cases, while the additional computations required for controlled steering contribute a very low overhead. The y-axis is shown on a log scale; the breakdown is nearly \textbf{identical across tasks}, indicating consistent per-timestep overhead.}
  \label{fig:task_computation}
\end{figure*}

\begin{figure*}[t]
  \centering
  \includegraphics[width=\textwidth]{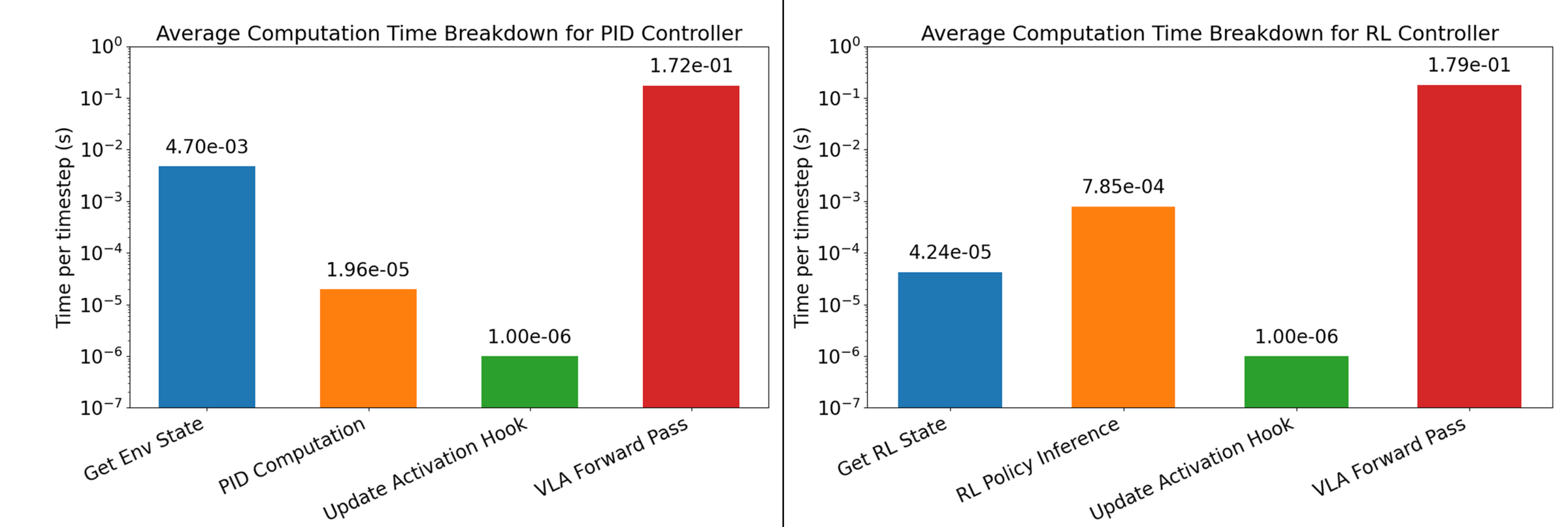}
  \caption{ Breakdown of the \textbf{average} per-timestep computational cost for CTRL-STEER with only PID (left) and RL+PID (right). The VLA forward pass dominates the runtime in both cases, while the additional computations required for controlled steering contribute a very low overhead.}
  \label{fig:avgcomputation}
\end{figure*}

\begin{figure*}[t]
  \centering
  \includegraphics[width=\textwidth]{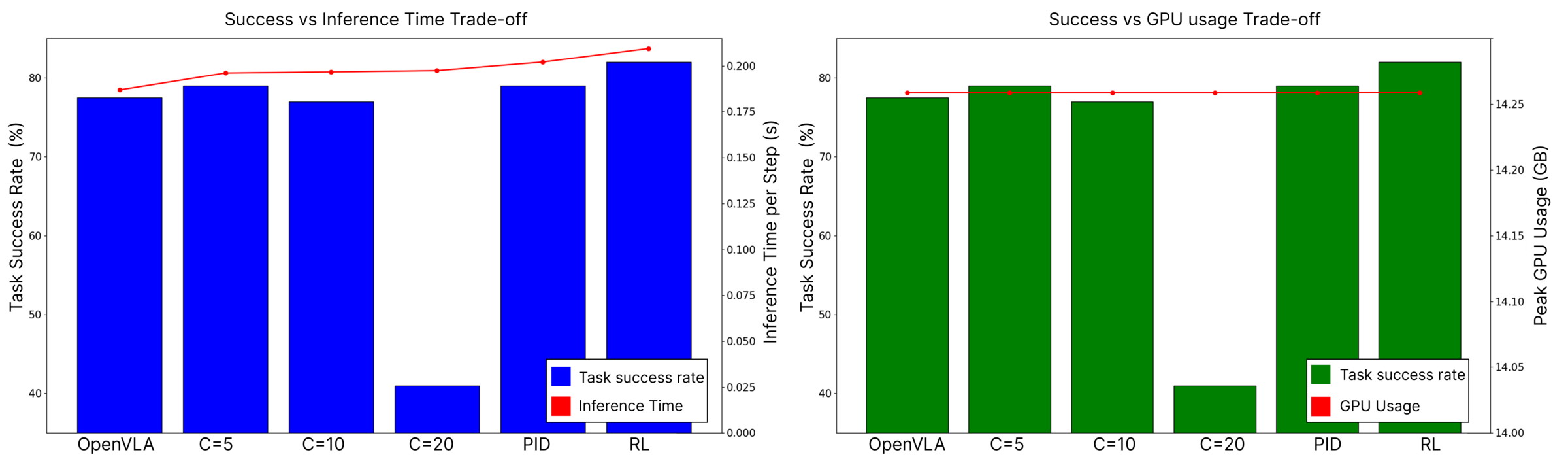}
  \caption{Trade-off between task success and computational cost across steering methods. Bars indicate task success rate, while the red line shows the per-timestep inference time (left) and peak GPU memory usage (right). CTRL-STEER (PID and RL+PID) achieves the highest success rates with only a modest increase in computational overhead compared to static steering baselines.}
  \label{fig:paretocomputation}
\end{figure*}

\cref{fig:task_computation} and \cref{fig:avgcomputation} shows the breakdown of the time overhead in controlled steering. The results show that the major computational cost comes from VLA forward pass, and the overhead created by the controller is negligible. \cref{fig:paretocomputation} shows how the improvement in task success rate dominates over the increase in computational cost.

\begin{table*}[!th]
\centering
\caption{Average Inference Cost for All Approaches (No steering, Static steering with $C = 5,10,20$, and CTRL-STEER with PID and RL)}
\small
\begin{tabular}{lcc}
\toprule
Task & time / step (s) & Peak GPU (GB) \\
\midrule
OpenVLA & 0.1869 $\pm$ 0.0039 & 14.2587 $\pm$ 0.0003 \\
C=5 & 0.1961 $\pm$ 0.0124 & 14.2587 $\pm$ 0.0003 \\
C=10 & 0.1967 $\pm$ 0.0079 & 14.2587 $\pm$ 0.0003 \\
C=20 & 0.1974 $\pm$ 0.0060 & 14.2587 $\pm$ 0.0003 \\
CTRL-STEER (PID) & 0.2021 $\pm$ 0.0076 & 14.2587 $\pm$ 0.0003 \\
CTRL-STEER (RL) & 0.2094 $\pm$ 0.0077 & 14.2588 $\pm$ 0.0003 \\
\bottomrule
\end{tabular}
\label{tab:complexity_main}
\end{table*}

\begin{table*}[t]
\centering
\caption{Inference cost per timestep for CTRL-STEER (RL+PID) }
\small
\begin{tabular}{lcc}
\toprule
Task & time / step (s) & Peak GPU (GB) \\
\midrule
open the middle drawer of the cabinet & 0.2205 & 14.2587 \\
put the bowl on the stove & 0.2217 & 14.2587 \\
put the wine bottle on top of the cabinet & 0.2166 & 14.2590 \\
open the top drawer and put the bowl inside & 0.2083 & 14.2592 \\
put the bowl on top of the cabinet & 0.2098 & 14.2589 \\
push the plate to the front of the stove & 0.2070 & 14.2590 \\
put the cream cheese in the bowl & 0.2073 & 14.2590 \\
turn on the stove & 0.2043 & 14.2582 \\
put the bowl on the plate & 0.1956 & 14.2586 \\
put the wine bottle on the rack & 0.2031 & 14.2589 \\
\midrule
Average & 0.2094 $\pm$ 0.0077 & 14.2588 $\pm$ 0.0003 \\
\bottomrule
\end{tabular}
\label{tab:complexity_rl}
\end{table*}

\begin{table*}[t]
\centering
\caption{Inference cost per timestep for CTRL-STEER with only PID Steering}
\small
\begin{tabular}{lcc}
\toprule
Task & time / step (s) & Peak GPU (GB) \\
\midrule
open the middle drawer of the cabinet & 0.2158 & 14.2586 \\
put the bowl on the stove & 0.2155 & 14.2586 \\
put the wine bottle on top of the cabinet & 0.2047 & 14.2589 \\
open the top drawer and put the bowl inside & 0.2036 & 14.2591 \\
put the bowl on top of the cabinet & 0.1967 & 14.2587 \\
push the plate to the front of the stove & 0.1961 & 14.2589 \\
put the cream cheese in the bowl & 0.2015 & 14.2589 \\
turn on the stove & 0.1977 & 14.2581 \\
put the bowl on the plate & 0.1956 & 14.2584 \\
put the wine bottle on the rack & 0.1933 & 14.2587 \\
\midrule
Average & 0.2021 $\pm$ 0.0076 & 14.2587 $\pm$ 0.0003 \\
\bottomrule
\end{tabular}
\label{tab:complexity_pid}
\end{table*}

\begin{table*}[t]
\centering
\caption{Inference cost per timestep for static steering with C=5~\cite{haon2025mechanistic} }
\small
\begin{tabular}{lcc}
\toprule
Task & time / step (s) & Peak GPU (GB) \\
\midrule
open the middle drawer of the cabinet & 0.2243 & 14.2586 \\
put the bowl on the stove & 0.2135 & 14.2586 \\
put the wine bottle on top of the cabinet & 0.1928 & 14.2589 \\
open the top drawer and put the bowl inside & 0.1903 & 14.2591 \\
put the bowl on top of the cabinet & 0.1947 & 14.2587 \\
push the plate to the front of the stove & 0.1876 & 14.2589 \\
put the cream cheese in the bowl & 0.1994 & 14.2589 \\
turn on the stove & 0.1871 & 14.2581 \\
put the bowl on the plate & 0.1882 & 14.2584 \\
put the wine bottle on the rack & 0.1833 & 14.2587 \\
\midrule
Average & 0.1961 $\pm$ 0.0124 & 14.2587 $\pm$ 0.0003 \\
\bottomrule
\end{tabular}
\label{tab:complexity_c5}
\end{table*}

\begin{table*}[t]
\centering
\caption{Inference cost per timestep for static steering with C=10~\cite{haon2025mechanistic}  }
\small
\begin{tabular}{lcc}
\toprule
Task & time / step (s) & Peak GPU (GB) \\
\midrule
open the middle drawer of the cabinet & 0.2147 & 14.2586 \\
put the bowl on the stove & 0.2081 & 14.2586 \\
put the wine bottle on top of the cabinet & 0.1964 & 14.2589 \\
open the top drawer and put the bowl inside & 0.1936 & 14.2591 \\
put the bowl on top of the cabinet & 0.1927 & 14.2587 \\
push the plate to the front of the stove & 0.1891 & 14.2589 \\
put the cream cheese in the bowl & 0.1976 & 14.2589 \\
turn on the stove & 0.1898 & 14.2581 \\
put the bowl on the plate & 0.1943 & 14.2584 \\
put the wine bottle on the rack & 0.1909 & 14.2587 \\
\midrule
Average & 0.1967 $\pm$ 0.0079 & 14.2587 $\pm$ 0.0003 \\
\bottomrule
\end{tabular}
\label{tab:complexity_c10}
\end{table*}

\begin{table*}[t]
\centering
\caption{Inference cost per timestep for static steering with C=20~\cite{haon2025mechanistic}  }
\small
\begin{tabular}{lcc}
\toprule
Task & time / step (s) & Peak GPU (GB) \\
\midrule
open the middle drawer of the cabinet & 0.2125 & 14.2586 \\
put the bowl on the stove & 0.2033 & 14.2586 \\
put the wine bottle on top of the cabinet & 0.1995 & 14.2589 \\
open the top drawer and put the bowl inside & 0.1939 & 14.2591 \\
put the bowl on top of the cabinet & 0.1935 & 14.2587 \\
push the plate to the front of the stove & 0.1935 & 14.2589 \\
put the cream cheese in the bowl & 0.1937 & 14.2589 \\
turn on the stove & 0.1921 & 14.2581 \\
put the bowl on the plate & 0.1973 & 14.2584 \\
put the wine bottle on the rack & 0.1950 & 14.2587 \\
\midrule
Average & 0.1974 $\pm$ 0.0060 & 14.2587 $\pm$ 0.0003 \\
\bottomrule
\end{tabular}
\label{tab:complexity_c20}
\end{table*}

\begin{table*}[t]
\centering
\caption{Inference cost per timestep for unsteered OpenVLA model}
\small
\begin{tabular}{lcc}
\toprule
Task & time / step (s) & Peak GPU (GB) \\
\midrule
open the middle drawer of the cabinet & 0.1954 & 14.2586 \\
put the bowl on the stove & 0.1900 & 14.2586 \\
put the wine bottle on top of the cabinet & 0.1835 & 14.2589 \\
open the top drawer and put the bowl inside & 0.1869 & 14.2591 \\
put the bowl on top of the cabinet & 0.1842 & 14.2587 \\
push the plate to the front of the stove & 0.1844 & 14.2589 \\
put the cream cheese in the bowl & 0.1915 & 14.2589 \\
turn on the stove & 0.1865 & 14.2581 \\
put the bowl on the plate & 0.1840 & 14.2584 \\
put the wine bottle on the rack & 0.1825 & 14.2587 \\
\midrule
Average & 0.1869 $\pm$ 0.0039 & 14.2587 $\pm$ 0.0003 \\
\bottomrule
\end{tabular}
\label{tab:complexity_naive}
\end{table*}

\section{Task-wise results}
\label{sec:takwise}

\cref{tab:long_height} to \cref{tab:spatial_speed} show the task-wise results for our experiments on each task suite in LIBERO. We compare our results with the unsteered OpenVLA model~\cite{kim2024openvla}, static steered model~\cite{haon2025mechanistic}, and our CTRL-STEER steering tool with both only PID controller and RL+PID controller (RL training initialized with PID data). For Static steering, for each task we chose the $\alpha$ value that gave the best result. For each task, $20$ individual rollouts were run deterministically by setting seed values.

\begin{table*}[t]
\centering
\caption{Detailed task-wise results on Height intervention for LIBERO LONG tasksuite}
\small
\scalebox{0.9}{
\begin{tabular}{llcccc}
\toprule
Task & Method & Height (m) & 95th percentile & AAT & SR (\%) \\
\midrule
\multirow{4}{*}{\makecell[l]{put both the alphabet \\ soup and the tomato \\ sauce in the basket}} & OpenVLA\cite{kim2024openvla} & 0.60 $\pm$ 0.02 & 0.70 $\pm$ 0.01 & 127.22 $\pm$ 30.18 & 65.00 \\
 & STATIC\cite{haon2025mechanistic} & 0.62 $\pm$ 0.01 & 0.71 $\pm$ 0.02 & 137.82 $\pm$ 36.92 & 65.00 \\
 & Ours (PID) & 0.62 $\pm$ 0.01 & 0.70 $\pm$ 0.02 & 125.25 $\pm$ 30.22 & 55.00 \\
 & Ours (RL) & 0.61 $\pm$ 0.01 & 0.71 $\pm$ 0.01 & 130.40 $\pm$ 28.37 & 65.00 \\
\midrule
\multirow{4}{*}{\makecell[l]{put the cheese box \\and the butter in the \\ basket}} & OpenVLA\cite{kim2024openvla} & 0.60 $\pm$ 0.01 & 0.70 $\pm$ 0.01 & 94.51 $\pm$ 9.98 & 75.00 \\
 & STATIC\cite{haon2025mechanistic} & 0.60 $\pm$ 0.03 & 0.70 $\pm$ 0.01 & 108.97 $\pm$ 41.47 & 80.00 \\
 & Ours (PID) & 0.60 $\pm$ 0.02 & 0.70 $\pm$ 0.01 & 88.59 $\pm$ 9.81 & 85.00 \\
 & Ours (RL) & 0.59 $\pm$ 0.02 & 0.70 $\pm$ 0.02 & 92.74 $\pm$ 18.01 & 65.00 \\
\midrule
\multirow{4}{*}{\makecell[l]{Turn on the stove and \\ put the moka \\ pot on it}} & OpenVLA\cite{kim2024openvla} & 1.04 $\pm$ 0.02 & 1.17 $\pm$ 0.01 & 233.30 $\pm$ 32.17 & 70.00 \\
 & STATIC\cite{haon2025mechanistic} & 1.04 $\pm$ 0.01 & 1.17 $\pm$ 0.01 & 227.19 $\pm$ 28.99 & 50.00 \\
 & Ours (PID) & 1.04 $\pm$ 0.02 & 1.17 $\pm$ 0.00 & 268.83 $\pm$ 153.20 & 60.00 \\
 & Ours (RL) & 1.02 $\pm$ 0.03 & 1.16 $\pm$ 0.02 & 297.11 $\pm$ 127.66 & 40.00 \\
\midrule
\multirow{4}{*}{\makecell[l]{put the black bowl in \\ the bottom drawer of \\ the  cabinet and close it}} & OpenVLA\cite{kim2024openvla} & 1.03 $\pm$ 0.03 & 1.16 $\pm$ 0.02 & 233.57 $\pm$ 82.29 & 50.00 \\
 & STATIC\cite{haon2025mechanistic} & 1.04 $\pm$ 0.04 & 1.17 $\pm$ 0.01 & 264.11 $\pm$ 149.17 & 50.00 \\
 & Ours (PID) & 1.01 $\pm$ 0.02 & 1.16 $\pm$ 0.01 & 292.53 $\pm$ 131.56 & 45.00 \\
 & Ours (RL) & 1.02 $\pm$ 0.03 & 1.16 $\pm$ 0.02 & 268.07 $\pm$ 96.26 & 45.00 \\
\midrule
\multirow{4}{*}{\makecell[l]{put the white mug \\on the left plate and \\ put the yellow and white \\mug on the right plate}} & OpenVLA\cite{kim2024openvla} & 0.59 $\pm$ 0.01 & 0.67 $\pm$ 0.01 & 106.93 $\pm$ 13.39 & 40.00 \\
 & STATIC\cite{haon2025mechanistic} & 0.59 $\pm$ 0.02 & 0.66 $\pm$ 0.02 & 117.07 $\pm$ 38.99 & 55.00 \\
 & Ours (PID) & 0.59 $\pm$ 0.02 & 0.66 $\pm$ 0.02 & 114.30 $\pm$ 36.05 & 35.00 \\
 & Ours (RL) & 0.59 $\pm$ 0.02 & 0.67 $\pm$ 0.01 & 124.15 $\pm$ 57.65 & 65.00 \\
\midrule
\multirow{4}{*}{\makecell[l]{pick up the book and \\place it in the back \\compartment of the caddy}} & OpenVLA\cite{kim2024openvla} & 1.16 $\pm$ 0.01 & 1.28 $\pm$ 0.03 & 172.27 $\pm$ 43.00 & 75.00 \\
 & STATIC\cite{haon2025mechanistic} & 1.16 $\pm$ 0.01 & 1.27 $\pm$ 0.02 & 185.08 $\pm$ 44.59 & 80.00 \\
 & Ours (PID) & 1.16 $\pm$ 0.01 & 1.28 $\pm$ 0.03 & 179.58 $\pm$ 38.63 & 80.00 \\
 & Ours (RL) & 1.16 $\pm$ 0.02 & 1.28 $\pm$ 0.03 & 177.97 $\pm$ 27.15 & 80.00 \\
\midrule
\multirow{4}{*}{\makecell[l]{put the white mug \\on the plate and \\put the chocolate pudding\\ to the right of the plate}} & OpenVLA\cite{kim2024openvla} & 0.55 $\pm$ 0.03 & 0.67 $\pm$ 0.03 & 88.62 $\pm$ 20.16 & 50.00 \\
 & STATIC\cite{haon2025mechanistic} & 0.55 $\pm$ 0.01 & 0.67 $\pm$ 0.02 & 90.82 $\pm$ 34.13 & 55.00 \\
 & Ours (PID) & 0.55 $\pm$ 0.03 & 0.67 $\pm$ 0.02 & 94.31 $\pm$ 42.52 & 50.00 \\
 & Ours (RL) & 0.55 $\pm$ 0.02 & 0.67 $\pm$ 0.01 & 85.34 $\pm$ 23.64 & 75.00 \\
\midrule
\multirow{4}{*}{\makecell[l]{put both the alphabet soup \\and the cream cheese \\box in the basket}} & OpenVLA\cite{kim2024openvla} & 0.60 $\pm$ 0.01 & 0.69 $\pm$ 0.01 & 113.02 $\pm$ 32.90 & 65.00 \\
 & STATIC\cite{haon2025mechanistic} & 0.59 $\pm$ 0.02 & 0.69 $\pm$ 0.01 & 100.24 $\pm$ 15.66 & 55.00 \\
 & Ours (PID) & 0.59 $\pm$ 0.02 & 0.69 $\pm$ 0.01 & 104.71 $\pm$ 26.79 & 55.00 \\
 & Ours (RL) & 0.59 $\pm$ 0.03 & 0.69 $\pm$ 0.01 & 108.56 $\pm$ 29.67 & 50.00 \\
\midrule
\multirow{4}{*}{\makecell[l]{put both moka pots \\on the stove}} & OpenVLA\cite{kim2024openvla} & 1.10 $\pm$ 0.01 & 1.19 $\pm$ 0.01 & 406.09 $\pm$ 79.07 & 40.00 \\
 & STATIC\cite{haon2025mechanistic} & 1.09 $\pm$ 0.01 & 1.18 $\pm$ 0.01 & 433.88 $\pm$ 86.87 & 45.00 \\
 & Ours (PID) & 1.09 $\pm$ 0.02 & 1.18 $\pm$ 0.02 & 417.96 $\pm$ 81.50 & 25.00 \\
 & Ours (RL) & 1.09 $\pm$ 0.01 & 1.18 $\pm$ 0.01 & 451.37 $\pm$ 155.28 & 35.00 \\
\midrule
\multirow{4}{*}{\makecell[l]{put the yellow and\\ white mug in the\\ microwave and close it}} & OpenVLA\cite{kim2024openvla} & 1.06 $\pm$ 0.03 & 1.17 $\pm$ 0.01 & 335.49 $\pm$ 92.70 & 50.00 \\
 & STATIC\cite{haon2025mechanistic} & 1.07 $\pm$ 0.02 & 1.18 $\pm$ 0.01 & 306.57 $\pm$ 92.76 & 50.00 \\
 & Ours (PID) & 1.07 $\pm$ 0.02 & 1.18 $\pm$ 0.01 & 332.52 $\pm$ 156.19 & 50.00 \\
 & Ours (RL) & 1.07 $\pm$ 0.02 & 1.18 $\pm$ 0.02 & 305.47 $\pm$ 52.77 & 50.00 \\
\midrule
\bottomrule
\end{tabular}
}

\label{tab:long_height}
\end{table*}

\begin{table*}[t]
\centering
\small
\caption{Detailed task-wise results on Height intervention for LIBERO GOAL tasksuite}
\scalebox{0.9}{
\begin{tabular}{llcccc}
\toprule
Task & Method & Height (m) & 95th percentile & AAT & SR (\%) \\
\midrule
\multirow{4}{*}{\makecell[l]{Open the middle layer\\ of the drawer}} & OpenVLA\cite{kim2024openvla} & 1.08 $\pm$ 0.03 & 1.16 $\pm$ 0.02 & 183.97 $\pm$ 137.46 & 60.00 \\
 & STATIC\cite{haon2025mechanistic} & 1.08 $\pm$ 0.02 & 1.16 $\pm$ 0.01 & 175.86 $\pm$ 119.38 & 70.00 \\
 & Ours (PID) & 1.08 $\pm$ 0.02 & 1.16 $\pm$ 0.02 & 201.36 $\pm$ 163.33 & 70.00 \\
 & Ours (RL) & 1.09 $\pm$ 0.02 & 1.17 $\pm$ 0.01 & 137.33 $\pm$ 58.08 & 55.00 \\
\midrule
\multirow{4}{*}{\makecell[l]{Put the bowl on\\ the stove}} & OpenVLA\cite{kim2024openvla} & 1.04 $\pm$ 0.03 & 1.17 $\pm$ 0.01 & 88.07 $\pm$ 82.83 & 100.00 \\
 & STATIC\cite{haon2025mechanistic} & 1.04 $\pm$ 0.02 & 1.17 $\pm$ 0.01 & 70.46 $\pm$ 13.55 & 100.00 \\
 & Ours (PID) & 1.04 $\pm$ 0.02 & 1.17 $\pm$ 0.01 & 72.83 $\pm$ 15.09 & 95.00 \\
 & Ours (RL) & 1.04 $\pm$ 0.03 & 1.17 $\pm$ 0.01 & 82.37 $\pm$ 57.13 & 95.00 \\
\midrule
\multirow{4}{*}{\makecell[l]{Put the wine bottle\\ on the top of the drawer}} & OpenVLA\cite{kim2024openvla} & 1.15 $\pm$ 0.02 & 1.30 $\pm$ 0.03 & 92.33 $\pm$ 44.63 & 80.00 \\
 & STATIC\cite{haon2025mechanistic} & 1.14 $\pm$ 0.02 & 1.30 $\pm$ 0.02 & 87.87 $\pm$ 18.78 & 75.00 \\
 & Ours (PID) & 1.14 $\pm$ 0.02 & 1.29 $\pm$ 0.04 & 140.67 $\pm$ 169.85 & 70.00 \\
 & Ours (RL) & 1.14 $\pm$ 0.03 & 1.30 $\pm$ 0.03 & 107.73 $\pm$ 68.19 & 85.00 \\
\midrule
\multirow{4}{*}{\makecell[l]{Open the top layer\\ of the drawer and put \\ the bowl inside}} & OpenVLA\cite{kim2024openvla} & 1.09 $\pm$ 0.02 & 1.21 $\pm$ 0.03 & 162.55 $\pm$ 23.14 & 50.00 \\
 & STATIC\cite{haon2025mechanistic} & 1.10 $\pm$ 0.02 & 1.22 $\pm$ 0.01 & 153.40 $\pm$ 14.61 & 60.00 \\
 & Ours (PID) & 1.10 $\pm$ 0.02 & 1.22 $\pm$ 0.03 & 210.72 $\pm$ 130.48 & 40.00 \\
 & Ours (RL) & 1.10 $\pm$ 0.01 & 1.22 $\pm$ 0.02 & 147.53 $\pm$ 14.90 & 65.00 \\
\midrule
\multirow{4}{*}{\makecell[l]{Put the bowl on \\the top of the drawer}} & OpenVLA\cite{kim2024openvla} & 1.08 $\pm$ 0.02 & 1.23 $\pm$ 0.02 & 74.13 $\pm$ 13.33 & 75.00 \\
 & STATIC\cite{haon2025mechanistic} & 1.08 $\pm$ 0.02 & 1.22 $\pm$ 0.02 & 74.34 $\pm$ 14.23 & 95.00 \\
 & Ours (PID) & 1.08 $\pm$ 0.02 & 1.22 $\pm$ 0.02 & 76.09 $\pm$ 18.18 & 100.00 \\
 & Ours (RL) & 1.08 $\pm$ 0.02 & 1.22 $\pm$ 0.02 & 82.13 $\pm$ 40.05 & 95.00 \\
\midrule
\multirow{4}{*}{\makecell[l]{Push the plate to\\ the front of the stove}} & OpenVLA\cite{kim2024openvla} & 0.98 $\pm$ 0.03 & 1.16 $\pm$ 0.02 & 109.82 $\pm$ 36.82 & 80.00 \\
 & STATIC\cite{haon2025mechanistic} & 0.97 $\pm$ 0.01 & 1.16 $\pm$ 0.02 & 111.36 $\pm$ 39.18 & 70.00 \\
 & Ours (PID) & 0.98 $\pm$ 0.02 & 1.17 $\pm$ 0.02 & 105.14 $\pm$ 16.00 & 85.00 \\
 & Ours (RL) & 0.98 $\pm$ 0.03 & 1.16 $\pm$ 0.02 & 107.90 $\pm$ 27.97 & 90.00 \\
\midrule
\multirow{4}{*}{\makecell[l]{Put the cream cheese\\ on the bowl}} & OpenVLA\cite{kim2024openvla} & 1.04 $\pm$ 0.02 & 1.17 $\pm$ 0.02 & 97.46 $\pm$ 42.33 & 75.00 \\
 & STATIC\cite{haon2025mechanistic} & 1.04 $\pm$ 0.03 & 1.17 $\pm$ 0.01 & 107.59 $\pm$ 121.07 & 90.00 \\
 & Ours (PID) & 1.05 $\pm$ 0.03 & 1.17 $\pm$ 0.01 & 94.57 $\pm$ 39.71 & 85.00 \\
 & Ours (RL) & 1.04 $\pm$ 0.03 & 1.17 $\pm$ 0.01 & 82.33 $\pm$ 25.66 & 70.00 \\
\midrule
\multirow{4}{*}{\makecell[l]{Turn on the stove}} & OpenVLA\cite{kim2024openvla} & 1.03 $\pm$ 0.03 & 1.16 $\pm$ 0.05 & 89.25 $\pm$ 100.89 & 95.00 \\
 & STATIC\cite{haon2025mechanistic} & 1.04 $\pm$ 0.03 & 1.18 $\pm$ 0.01 & 82.72 $\pm$ 82.80 & 95.00 \\
 & Ours (PID) & 1.03 $\pm$ 0.03 & 1.18 $\pm$ 0.01 & 76.68 $\pm$ 40.95 & 90.00 \\
 & Ours (RL) & 1.04 $\pm$ 0.03 & 1.18 $\pm$ 0.01 & 82.20 $\pm$ 79.16 & 100.00 \\
\midrule
\multirow{4}{*}{\makecell[l]{Put the bowl on \\the plate}} & OpenVLA\cite{kim2024openvla} & 1.02 $\pm$ 0.01 & 1.17 $\pm$ 0.01 & 57.11 $\pm$ 4.56 & 95.00 \\
 & STATIC\cite{haon2025mechanistic} & 1.02 $\pm$ 0.01 & 1.16 $\pm$ 0.02 & 75.93 $\pm$ 67.10 & 100.00 \\
 & Ours (PID) & 1.03 $\pm$ 0.01 & 1.16 $\pm$ 0.01 & 62.36 $\pm$ 10.42 & 95.00 \\
 & Ours (RL) & 1.03 $\pm$ 0.02 & 1.17 $\pm$ 0.01 & 66.07 $\pm$ 26.61 & 90.00 \\
\midrule
\multirow{4}{*}{\makecell[l]{Put the wine bottle \\on the rack}} & OpenVLA\cite{kim2024openvla} & 1.09 $\pm$ 0.05 & 1.23 $\pm$ 0.03 & 193.83 $\pm$ 152.36 & 65.00 \\
 & STATIC\cite{haon2025mechanistic} & 1.11 $\pm$ 0.02 & 1.25 $\pm$ 0.02 & 135.47 $\pm$ 62.08 & 65.00 \\
 & Ours (PID) & 1.11 $\pm$ 0.02 & 1.24 $\pm$ 0.03 & 154.05 $\pm$ 124.58 & 60.00 \\
 & Ours (RL) & 1.11 $\pm$ 0.02 & 1.24 $\pm$ 0.02 & 136.68 $\pm$ 95.22 & 75.00 \\
\midrule
\bottomrule
\end{tabular}
}

\label{tab:goal_height}
\end{table*}

\begin{table*}[t]
\centering
\caption{Detailed task-wise results on Height intervention for LIBERO OBJECT tasksuite}
\small
\scalebox{0.9}{
\begin{tabular}{llcccc}
\toprule
Task & Method & Height (m) & 95th percentile & AAT & SR (\%) \\
\midrule
\multirow{4}{*}{\makecell[l]{Pick the alphabet\\ soup and place it \\in the basket}} & OpenVLA\cite{kim2024openvla} & 0.22 $\pm$ 0.01 & 0.30 $\pm$ 0.02 & 2.48 $\pm$ 1.03 & 90.00 \\
 & STATIC\cite{haon2025mechanistic} & 0.22 $\pm$ 0.02 & 0.30 $\pm$ 0.02 & 2.85 $\pm$ 1.77 & 85.00 \\
 & Ours (PID) & 0.21 $\pm$ 0.01 & 0.30 $\pm$ 0.02 & 2.42 $\pm$ 0.95 & 80.00 \\
 & Ours (RL) & 0.21 $\pm$ 0.02 & 0.29 $\pm$ 0.01 & 2.76 $\pm$ 0.86 & 85.00 \\
\midrule
\multirow{4}{*}{\makecell[l]{Pick up the cream \\cheese and \\ place it \\in the  basket}} & OpenVLA\cite{kim2024openvla} & 0.17 $\pm$ 0.01 & 0.26 $\pm$ 0.01 & 0.18 $\pm$ 0.13 & 70.00 \\
 & STATIC\cite{haon2025mechanistic} & 0.17 $\pm$ 0.04 & 0.26 $\pm$ 0.01 & 0.17 $\pm$ 0.20 & 55.00 \\
 & Ours (PID) & 0.17 $\pm$ 0.01 & 0.26 $\pm$ 0.01 & 0.19 $\pm$ 0.11 & 85.00 \\
 & Ours (RL) & 0.16 $\pm$ 0.02 & 0.26 $\pm$ 0.01 & 0.15 $\pm$ 0.09 & 75.00 \\
\midrule
\multirow{4}{*}{\makecell[l]{Pick up the salad\\ dressing and \\ place it \\in the basket}} & OpenVLA\cite{kim2024openvla} & 0.23 $\pm$ 0.02 & 0.31 $\pm$ 0.02 & 6.14 $\pm$ 2.28 & 75.00 \\
 & STATIC\cite{haon2025mechanistic} & 0.23 $\pm$ 0.01 & 0.31 $\pm$ 0.01 & 6.38 $\pm$ 2.05 & 90.00 \\
 & Ours (PID) & 0.23 $\pm$ 0.01 & 0.32 $\pm$ 0.01 & 6.21 $\pm$ 2.30 & 80.00 \\
 & Ours (RL) & 0.23 $\pm$ 0.01 & 0.32 $\pm$ 0.01 & 5.52 $\pm$ 0.98 & 80.00 \\
\midrule
\multirow{4}{*}{\makecell[l]{Pick up the bbq \\sauce and  place it in \\the  basket}} & OpenVLA\cite{kim2024openvla} & 0.19 $\pm$ 0.04 & 0.28 $\pm$ 0.02 & 2.40 $\pm$ 1.42 & 40.00 \\
 & STATIC\cite{haon2025mechanistic} & 0.22 $\pm$ 0.01 & 0.28 $\pm$ 0.01 & 5.38 $\pm$ 1.56 & 45.00 \\
 & Ours (PID) & 0.21 $\pm$ 0.03 & 0.29 $\pm$ 0.02 & 10.01 $\pm$ 17.24 & 50.00 \\
 & Ours (RL) & 0.21 $\pm$ 0.03 & 0.29 $\pm$ 0.03 & 7.04 $\pm$ 8.63 & 50.00 \\
\midrule
\multirow{4}{*}{\makecell[l]{Pick up the ketchup and \\ place it in the \\ basket}} & OpenVLA\cite{kim2024openvla} & 0.24 $\pm$ 0.01 & 0.31 $\pm$ 0.02 & 8.19 $\pm$ 1.86 & 90.00 \\
 & STATIC\cite{haon2025mechanistic} & 0.24 $\pm$ 0.01 & 0.31 $\pm$ 0.02 & 13.69 $\pm$ 12.11 & 80.00 \\
 & Ours (PID) & 0.24 $\pm$ 0.01 & 0.32 $\pm$ 0.03 & 7.60 $\pm$ 1.12 & 85.00 \\
 & Ours (RL) & 0.24 $\pm$ 0.02 & 0.31 $\pm$ 0.02 & 13.33 $\pm$ 12.71 & 80.00 \\
\midrule
\multirow{4}{*}{\makecell[l]{Pick up the tomato\\ sauce and  place it \\in the basket}} & OpenVLA\cite{kim2024openvla} & 0.19 $\pm$ 0.02 & 0.27 $\pm$ 0.01 & 3.56 $\pm$ 1.50 & 75.00 \\
 & STATIC\cite{haon2025mechanistic} & 0.20 $\pm$ 0.01 & 0.27 $\pm$ 0.01 & 3.15 $\pm$ 0.91 & 75.00 \\
 & Ours (PID) & 0.17 $\pm$ 0.04 & 0.26 $\pm$ 0.01 & 3.52 $\pm$ 1.04 & 80.00 \\
 & Ours (RL) & 0.19 $\pm$ 0.02 & 0.26 $\pm$ 0.01 & 3.87 $\pm$ 1.42 & 80.00 \\
\midrule
\multirow{4}{*}{\makecell[l]{Pick up the butter \\and  place it in the \\ basket}} & OpenVLA\cite{kim2024openvla} & 0.20 $\pm$ 0.01 & 0.30 $\pm$ 0.02 & 0.15 $\pm$ 0.04 & 60.00 \\
 & STATIC\cite{haon2025mechanistic} & 0.19 $\pm$ 0.01 & 0.27 $\pm$ 0.02 & 0.36 $\pm$ 0.26 & 75.00 \\
 & Ours (PID) & 0.20 $\pm$ 0.01 & 0.29 $\pm$ 0.02 & 0.14 $\pm$ 0.05 & 60.00 \\
 & Ours (RL) & 0.20 $\pm$ 0.01 & 0.28 $\pm$ 0.02 & 0.20 $\pm$ 0.10 & 60.00 \\
\midrule
\multirow{4}{*}{\makecell[l]{Pick up the milk and \\ place it in the \\ basket}} & OpenVLA\cite{kim2024openvla} & 0.24 $\pm$ 0.01 & 0.32 $\pm$ 0.02 & 7.20 $\pm$ 5.11 & 70.00 \\
 & STATIC\cite{haon2025mechanistic} & 0.24 $\pm$ 0.01 & 0.32 $\pm$ 0.02 & 6.48 $\pm$ 2.11 & 90.00 \\
 & Ours (PID) & 0.24 $\pm$ 0.01 & 0.32 $\pm$ 0.02 & 6.35 $\pm$ 4.42 & 90.00 \\
 & Ours (RL) & 0.24 $\pm$ 0.01 & 0.32 $\pm$ 0.02 & 6.59 $\pm$ 3.29 & 85.00 \\
\midrule
\multirow{4}{*}{\makecell[l]{Pick up the pudding and \\ place it in the \\ basket}} & OpenVLA\cite{kim2024openvla} & 0.18 $\pm$ 0.04 & 0.28 $\pm$ 0.03 & 0.27 $\pm$ 0.38 & 60.00 \\
 & STATIC\cite{haon2025mechanistic} & 0.19 $\pm$ 0.01 & 0.28 $\pm$ 0.02 & 0.22 $\pm$ 0.11 & 85.00 \\
 & Ours (PID) & 0.19 $\pm$ 0.03 & 0.28 $\pm$ 0.02 & 0.20 $\pm$ 0.06 & 65.00 \\
 & Ours (RL) & 0.19 $\pm$ 0.02 & 0.28 $\pm$ 0.01 & 0.33 $\pm$ 0.22 & 85.00 \\
\midrule
\multirow{4}{*}{\makecell[l]{Pick up the orange\\ juice and  place it in\\ the basket}} & OpenVLA\cite{kim2024openvla} & 0.23 $\pm$ 0.02 & 0.29 $\pm$ 0.02 & 9.68 $\pm$ 9.08 & 90.00 \\
 & STATIC\cite{haon2025mechanistic} & 0.23 $\pm$ 0.01 & 0.31 $\pm$ 0.02 & 5.40 $\pm$ 2.56 & 95.00 \\
 & Ours (PID) & 0.23 $\pm$ 0.01 & 0.30 $\pm$ 0.02 & 8.11 $\pm$ 7.48 & 85.00 \\
 & Ours (RL) & 0.23 $\pm$ 0.01 & 0.30 $\pm$ 0.01 & 6.46 $\pm$ 1.91 & 90.00 \\
\midrule
\bottomrule
\end{tabular}
}

\label{tab:object_height}
\end{table*}

\begin{table*}[t]
\centering
\caption{Detailed task-wise results on Height intervention for LIBERO SPATIAL tasksuite}
\small
\scalebox{0.9}{
\begin{tabular}{llcccc}
\toprule
Task & Method & Height (m) & 95th percentile & AAT & SR (\%) \\
\midrule
\multirow{4}{*}{\makecell[l]{Pick the akita black bowl \\ between the plate and the \\ramekin and place it on\\ the plate}} & OpenVLA\cite{kim2024openvla} & 1.04 $\pm$ 0.02 & 1.18 $\pm$ 0.01 & 71.44 $\pm$ 16.47 & 75.00 \\
 & STATIC\cite{haon2025mechanistic} & 1.05 $\pm$ 0.03 & 1.19 $\pm$ 0.01 & 74.07 $\pm$ 33.02 & 90.00 \\
 & Ours (PID) & 1.04 $\pm$ 0.01 & 1.18 $\pm$ 0.01 & 77.09 $\pm$ 48.70 & 80.00 \\
 & Ours (RL) & 1.05 $\pm$ 0.03 & 1.18 $\pm$ 0.01 & 85.69 $\pm$ 67.64 & 90.00 \\
\midrule
\multirow{4}{*}{\makecell[l]{Pick the akita black bowl next\\ to the ramekin and place\\ it on the plate}} & OpenVLA\cite{kim2024openvla} & 1.05 $\pm$ 0.01 & 1.18 $\pm$ 0.01 & 94.74 $\pm$ 12.98 & 95.00 \\
 & STATIC\cite{haon2025mechanistic} & 1.06 $\pm$ 0.01 & 1.17 $\pm$ 0.01 & 88.47 $\pm$ 5.23 & 100.00 \\
 & Ours (PID) & 1.06 $\pm$ 0.01 & 1.18 $\pm$ 0.01 & 87.40 $\pm$ 5.92 & 90.00 \\
 & Ours (RL) & 1.05 $\pm$ 0.02 & 1.17 $\pm$ 0.02 & 119.13 $\pm$ 133.02 & 100.00 \\
\midrule
\multirow{4}{*}{\makecell[l]{Pick the akita black bowl from\\ table center and place it\\ on the plate}} & OpenVLA\cite{kim2024openvla} & 1.06 $\pm$ 0.03 & 1.18 $\pm$ 0.01 & 104.24 $\pm$ 46.49 & 95.00 \\
 & STATIC\cite{haon2025mechanistic} & 1.06 $\pm$ 0.05 & 1.17 $\pm$ 0.01 & 156.20 $\pm$ 201.48 & 85.00 \\
 & Ours (PID) & 1.05 $\pm$ 0.02 & 1.17 $\pm$ 0.01 & 86.48 $\pm$ 12.31 & 75.00 \\
 & Ours (RL) & 1.05 $\pm$ 0.04 & 1.18 $\pm$ 0.01 & 112.69 $\pm$ 94.86 & 80.00 \\
\midrule
\multirow{4}{*}{\makecell[l]{Pick the akita black bowl next \\to the cookies box and \\place it on the plate}} & OpenVLA\cite{kim2024openvla} & 1.05 $\pm$ 0.01 & 1.18 $\pm$ 0.01 & 70.37 $\pm$ 7.73 & 85.00 \\
 & STATIC\cite{haon2025mechanistic} & 1.05 $\pm$ 0.01 & 1.18 $\pm$ 0.01 & 67.69 $\pm$ 4.97 & 100.00 \\
 & Ours (PID) & 1.05 $\pm$ 0.01 & 1.18 $\pm$ 0.01 & 68.78 $\pm$ 6.31 & 95.00 \\
 & Ours (RL) & 1.06 $\pm$ 0.01 & 1.18 $\pm$ 0.01 & 69.95 $\pm$ 6.78 & 95.00 \\
\midrule
\multirow{4}{*}{\makecell[l]{Pick the akita black bowl in \\the top layer of the wooden \\cabinet and place it on\\ the plate}} & OpenVLA\cite{kim2024openvla} & 1.15 $\pm$ 0.01 & 1.26 $\pm$ 0.02 & 109.62 $\pm$ 8.53 & 70.00 \\
 & STATIC\cite{haon2025mechanistic} & 1.15 $\pm$ 0.01 & 1.26 $\pm$ 0.01 & 107.59 $\pm$ 7.21 & 75.00 \\
 & Ours (PID) & 1.14 $\pm$ 0.03 & 1.25 $\pm$ 0.02 & 137.42 $\pm$ 89.71 & 65.00 \\
 & Ours (RL) & 1.16 $\pm$ 0.01 & 1.26 $\pm$ 0.01 & 110.86 $\pm$ 10.59 & 75.00 \\
\midrule
\multirow{4}{*}{\makecell[l]{Pick the akita black bowl on\\ the ramekin and place it on\\ the plate}} & OpenVLA\cite{kim2024openvla} & 1.05 $\pm$ 0.02 & 1.17 $\pm$ 0.01 & 81.43 $\pm$ 22.47 & 40.00 \\
 & STATIC\cite{haon2025mechanistic} & 1.07 $\pm$ 0.01 & 1.17 $\pm$ 0.01 & 81.18 $\pm$ 10.50 & 65.00 \\
 & Ours (PID) & 1.06 $\pm$ 0.01 & 1.17 $\pm$ 0.01 & 74.63 $\pm$ 9.66 & 45.00 \\
 & Ours (RL) & 1.06 $\pm$ 0.01 & 1.17 $\pm$ 0.01 & 78.89 $\pm$ 14.08 & 50.00 \\
\midrule
\multirow{4}{*}{\makecell[l]{Pick the akita black bowl on\\ the cookies box and place \\it on the plate}} & OpenVLA\cite{kim2024openvla} & 1.04 $\pm$ 0.01 & 1.17 $\pm$ 0.01 & 85.38 $\pm$ 9.90 & 95.00 \\
 & STATIC\cite{haon2025mechanistic} & 1.05 $\pm$ 0.02 & 1.17 $\pm$ 0.01 & 96.71 $\pm$ 32.52 & 95.00 \\
 & Ours (PID) & 1.05 $\pm$ 0.01 & 1.17 $\pm$ 0.01 & 82.47 $\pm$ 7.35 & 85.00 \\
 & Ours (RL) & 1.04 $\pm$ 0.01 & 1.17 $\pm$ 0.01 & 87.63 $\pm$ 13.66 & 80.00 \\
\midrule
\multirow{4}{*}{\makecell[l]{Pick the akita black bowl\\ on the stove and place it\\ on the plate}} & OpenVLA\cite{kim2024openvla} & 1.04 $\pm$ 0.02 & 1.16 $\pm$ 0.01 & 106.71 $\pm$ 37.09 & 70.00 \\
 & STATIC\cite{haon2025mechanistic} & 1.05 $\pm$ 0.01 & 1.16 $\pm$ 0.01 & 97.01 $\pm$ 6.68 & 75.00 \\
 & Ours (PID) & 1.04 $\pm$ 0.02 & 1.16 $\pm$ 0.02 & 100.66 $\pm$ 14.54 & 80.00 \\
 & Ours (RL) & 1.05 $\pm$ 0.01 & 1.16 $\pm$ 0.01 & 95.56 $\pm$ 6.37 & 80.00 \\
\midrule
\multirow{4}{*}{\makecell[l]{Pick the akita black bowl\\ next to the plate and\\ place it on the plate}} & OpenVLA\cite{kim2024openvla} & 1.04 $\pm$ 0.01 & 1.17 $\pm$ 0.01 & 78.04 $\pm$ 7.39 & 90.00 \\
 & STATIC\cite{haon2025mechanistic} & 1.05 $\pm$ 0.03 & 1.17 $\pm$ 0.01 & 182.60 $\pm$ 209.87 & 85.00 \\
 & Ours (PID) & 1.05 $\pm$ 0.03 & 1.17 $\pm$ 0.01 & 121.24 $\pm$ 137.60 & 80.00 \\
 & Ours (RL) & 1.04 $\pm$ 0.02 & 1.17 $\pm$ 0.01 & 109.99 $\pm$ 97.70 & 80.00 \\
\midrule
\multirow{4}{*}{\makecell[l]{Pick the akita black bowl on \\the wooden cabinet and place \\it on the plate}} & OpenVLA\cite{kim2024openvla} & 1.19 $\pm$ 0.02 & 1.29 $\pm$ 0.02 & 110.09 $\pm$ 20.07 & 65.00 \\
 & STATIC\cite{haon2025mechanistic} & 1.18 $\pm$ 0.02 & 1.28 $\pm$ 0.02 & 152.80 $\pm$ 126.99 & 75.00 \\
 & Ours (PID) & 1.19 $\pm$ 0.01 & 1.28 $\pm$ 0.02 & 101.64 $\pm$ 6.08 & 55.00 \\
 & Ours (RL) & 1.19 $\pm$ 0.01 & 1.29 $\pm$ 0.01 & 105.96 $\pm$ 7.40 & 65.00 \\
\midrule
\bottomrule
\end{tabular}
}
\label{tab:spatial_height}
\end{table*}

\begin{table*}[t]
\centering
\caption{Detailed task-wise results on Speed intervention for LIBERO LONG tasksuite}
\small
\scalebox{0.9}{
\begin{tabular}{llccc}
\toprule
Task & Method & Speed (cm/s) & SAT & SR (\%) \\
\midrule
\multirow{4}{*}{\makecell[l]{put both the alphabet \\ soup and the tomato \\ sauce in the basket}} & OpenVLA\cite{kim2024openvla} & 11.79 $\pm$ 2.30 & 2.80 $\pm$ 0.33 & 65.00 \\
 & STATIC\cite{haon2025mechanistic} & 11.69 $\pm$ 2.77 & 2.60 $\pm$ 0.49 & 60.00 \\
 & Ours (PID) & 11.88 $\pm$ 1.95 & 2.61 $\pm$ 0.45 & 60.00 \\
 & Ours (RL) & 11.04 $\pm$ 2.14 & 2.77 $\pm$ 0.53 & 75.00 \\
\midrule
\multirow{4}{*}{\makecell[l]{put the cheese box \\and the butter in the \\ basket}} & OpenVLA\cite{kim2024openvla} & 15.17 $\pm$ 1.38 & 3.26 $\pm$ 0.26 & 75.00 \\
 & STATIC\cite{haon2025mechanistic} & 13.47 $\pm$ 3.07 & 3.09 $\pm$ 0.27 & 75.00 \\
 & Ours (PID) & 14.30 $\pm$ 3.35 & 3.08 $\pm$ 0.29 & 80.00 \\
 & Ours (RL) & 14.26 $\pm$ 2.66 & 2.99 $\pm$ 0.43 & 75.00 \\
\midrule
\multirow{4}{*}{\makecell[l]{turn on the stove and \\ put the moka \\ pot on it}} & OpenVLA\cite{kim2024openvla} & 9.56 $\pm$ 1.01 & 2.99 $\pm$ 0.58 & 70.00 \\
 & STATIC\cite{haon2025mechanistic} & 10.30 $\pm$ 0.58 & 2.93 $\pm$ 0.84 & 55.00 \\
 & Ours (PID) & 10.37 $\pm$ 1.10 & 3.27 $\pm$ 0.57 & 75.00 \\
 & Ours (RL) & 10.01 $\pm$ 1.17 & 2.90 $\pm$ 0.41 & 60.00 \\
\midrule
\multirow{4}{*}{\makecell[l]{put the black bowl in \\ the bottom drawer of \\ the  cabinet and close it}} & OpenVLA\cite{kim2024openvla} & 10.11 $\pm$ 2.15 & 1.77 $\pm$ 0.97 & 50.00 \\
 & STATIC\cite{haon2025mechanistic} & 10.81 $\pm$ 1.65 & 1.46 $\pm$ 0.59 & 65.00 \\
 & Ours (PID) & 10.37 $\pm$ 2.42 & 1.70 $\pm$ 0.90 & 60.00 \\
 & Ours (RL) & 9.14 $\pm$ 3.59 & 1.29 $\pm$ 0.85 & 70.00 \\
\midrule
\multirow{4}{*}{\makecell[l]{put the white mug \\on the left plate and \\ put the yellow and white \\mug on the right plate}} & OpenVLA\cite{kim2024openvla} & 11.91 $\pm$ 1.31 & 0.99 $\pm$ 0.28 & 40.00 \\
 & STATIC\cite{haon2025mechanistic} & 10.27 $\pm$ 3.15 & 1.09 $\pm$ 0.27 & 50.00 \\
 & Ours (PID) & 10.80 $\pm$ 2.65 & 1.04 $\pm$ 0.21 & 45.00 \\
 & Ours (RL) & 11.62 $\pm$ 2.94 & 1.12 $\pm$ 0.24 & 45.00 \\
\midrule

\multirow{4}{*}{\makecell[l]{pick up the book and \\place it in the back \\compartment of the caddy}} & OpenVLA\cite{kim2024openvla} & 11.15 $\pm$ 1.46 & 4.81 $\pm$ 0.98 & 75.00 \\
 & STATIC\cite{haon2025mechanistic} & 10.81 $\pm$ 2.11 & 5.35 $\pm$ 0.77 & 70.00 \\
 & Ours (PID) & 10.15 $\pm$ 2.68 & 4.88 $\pm$ 1.09 & 60.00 \\
 & Ours (RL) & 10.76 $\pm$ 1.64 & 4.53 $\pm$ 0.89 & 80.00 \\
\midrule
\multirow{4}{*}{\makecell[l]{put the white mug \\on the plate and \\put the chocolate pudding\\ to the right of the plate}} & OpenVLA\cite{kim2024openvla} & 10.21 $\pm$ 2.62 & 2.07 $\pm$ 0.38 & 50.00 \\
 & STATIC\cite{haon2025mechanistic} & 9.99 $\pm$ 3.31 & 2.06 $\pm$ 0.64 & 60.00 \\
 & Ours (PID) & 9.53 $\pm$ 2.82 & 2.03 $\pm$ 0.40 & 55.00 \\
 & Ours (RL) & 10.84 $\pm$ 1.99 & 2.35 $\pm$ 0.54 & 30.00 \\
\midrule
\multirow{4}{*}{\makecell[l]{put both the alphabet soup \\and the cream cheese \\box in the basket}} & OpenVLA\cite{kim2024openvla} & 12.45 $\pm$ 2.38 & 3.45 $\pm$ 0.55 & 65.00 \\
 & STATIC\cite{haon2025mechanistic} & 12.68 $\pm$ 1.76 & 3.22 $\pm$ 0.48 & 60.00 \\
 & Ours (PID) & 11.54 $\pm$ 3.62 & 3.44 $\pm$ 0.60 & 65.00 \\
 & Ours (RL) & 12.18 $\pm$ 1.90 & 3.37 $\pm$ 0.35 & 60.00 \\
\midrule
\multirow{4}{*}{\makecell[l]{put both moka pots \\on the stove}} & OpenVLA\cite{kim2024openvla} & 9.86 $\pm$ 1.80 & 1.08 $\pm$ 0.40 & 40.00 \\
 & STATIC\cite{haon2025mechanistic} & 9.32 $\pm$ 1.91 & 0.91 $\pm$ 0.35 & 40.00 \\
 & Ours (PID) & 9.45 $\pm$ 2.55 & 1.18 $\pm$ 0.31 & 30.00 \\
 & Ours (RL) & 9.07 $\pm$ 2.24 & 0.94 $\pm$ 0.38 & 40.00 \\
\midrule
\multirow{4}{*}{\makecell[l]{put the yellow and\\ white mug in the\\ microwave and close it}} & OpenVLA\cite{kim2024openvla} & 8.45 $\pm$ 2.45 & 1.27 $\pm$ 0.34 & 50.00 \\
 & STATIC\cite{haon2025mechanistic} & 8.00 $\pm$ 3.01 & 1.66 $\pm$ 1.03 & 20.00 \\
 & Ours (PID) & 8.65 $\pm$ 1.74 & 1.36 $\pm$ 1.37 & 65.00 \\
 & Ours (RL) & 9.75 $\pm$ 1.18 & 1.41 $\pm$ 1.40 & 45.00 \\
\midrule
\bottomrule
\end{tabular}
}
\label{tab:long_speed}
\end{table*}

\begin{table*}[t]
\centering
\caption{Detailed task-wise results on Speed intervention for LIBERO GOAL tasksuite}
\small
\scalebox{0.9}{
\begin{tabular}{llccc}
\toprule
Task & Method & Speed (cm/s) & SAT & SR (\%) \\
\midrule
\multirow{4}{*}{\makecell[l]{Open the middle layer\\ of the drawer}} & OpenVLA\cite{kim2024openvla} & 7.59 $\pm$ 2.44 & 1.68 $\pm$ 2.18 & 60.00 \\
 & STATIC\cite{haon2025mechanistic} & 8.50 $\pm$ 2.13 & 0.79 $\pm$ 1.73 & 80.00 \\
 & Ours (PID) & 7.12 $\pm$ 1.86 & 1.54 $\pm$ 1.94 & 60.00 \\
 & Ours (RL) & 8.20 $\pm$ 2.28 & 0.20 $\pm$ 0.57 & 75.00 \\
\midrule
\multirow{4}{*}{\makecell[l]{Put the bowl on\\ the stove}} & OpenVLA\cite{kim2024openvla} & 17.47 $\pm$ 3.66 & 3.40 $\pm$ 0.39 & 100.00 \\
 & STATIC\cite{haon2025mechanistic} & 17.95 $\pm$ 1.71 & 3.45 $\pm$ 0.39 & 95.00 \\
 & Ours (PID) & 18.47 $\pm$ 1.60 & 3.45 $\pm$ 0.35 & 95.00 \\
 & Ours (RL) & 18.03 $\pm$ 1.97 & 3.50 $\pm$ 0.58 & 90.00 \\
\midrule
\multirow{4}{*}{\makecell[l]{Put the wine bottle\\ on the top of the drawer}} & OpenVLA\cite{kim2024openvla} & 14.64 $\pm$ 3.40 & 5.82 $\pm$ 0.62 & 80.00 \\
 & STATIC\cite{haon2025mechanistic} & 14.30 $\pm$ 2.48 & 5.79 $\pm$ 0.79 & 80.00 \\
 & Ours (PID) & 12.84 $\pm$ 3.71 & 5.52 $\pm$ 0.56 & 75.00 \\
 & Ours (RL) & 13.87 $\pm$ 4.99 & 5.59 $\pm$ 0.45 & 85.00 \\
\midrule
\multirow{4}{*}{\makecell[l]{Open the top layer\\ of the drawer and put \\ the bowl inside}} & OpenVLA\cite{kim2024openvla} & 14.60 $\pm$ 1.64 & 2.79 $\pm$ 0.41 & 50.00 \\
 & STATIC\cite{haon2025mechanistic} & 15.64 $\pm$ 1.23 & 2.90 $\pm$ 0.28 & 50.00 \\
 & Ours (PID) & 12.97 $\pm$ 4.71 & 2.94 $\pm$ 0.36 & 45.00 \\
 & Ours (RL) & 14.26 $\pm$ 1.52 & 2.89 $\pm$ 0.46 & 55.00 \\
\midrule
\multirow{4}{*}{\makecell[l]{Put the bowl on \\the top of the drawer}} & OpenVLA\cite{kim2024openvla} & 18.63 $\pm$ 1.91 & 4.59 $\pm$ 0.22 & 75.00 \\
 & STATIC\cite{haon2025mechanistic} & 18.60 $\pm$ 1.97 & 4.79 $\pm$ 0.31 & 100.00 \\
 & Ours (PID) & 18.25 $\pm$ 4.20 & 4.75 $\pm$ 0.35 & 95.00 \\
 & Ours (RL) & 17.71 $\pm$ 3.96 & 4.66 $\pm$ 0.31 & 95.00 \\
\midrule
\multirow{4}{*}{\makecell[l]{Push the plate to\\ the front of the stove}} & OpenVLA\cite{kim2024openvla} & 11.43 $\pm$ 1.78 & 3.35 $\pm$ 0.39 & 80.00 \\
 & STATIC\cite{haon2025mechanistic} & 12.19 $\pm$ 1.01 & 3.12 $\pm$ 0.46 & 85.00 \\
 & Ours (PID) & 12.13 $\pm$ 1.11 & 3.21 $\pm$ 0.41 & 85.00 \\
 & Ours (RL) & 11.96 $\pm$ 1.47 & 3.23 $\pm$ 0.39 & 100.00 \\
\midrule
\multirow{4}{*}{\makecell[l]{Put the cream cheese\\ on the bowl}} & OpenVLA\cite{kim2024openvla} & 14.45 $\pm$ 3.71 & 3.44 $\pm$ 0.54 & 75.00 \\
 & STATIC\cite{haon2025mechanistic} & 14.60 $\pm$ 4.63 & 3.75 $\pm$ 0.36 & 50.00 \\
 & Ours (PID) & 12.82 $\pm$ 4.80 & 3.70 $\pm$ 0.52 & 60.00 \\
 & Ours (RL) & 15.18 $\pm$ 3.52 & 3.74 $\pm$ 0.39 & 75.00 \\
\midrule
\multirow{4}{*}{\makecell[l]{Turn on the stove}} & OpenVLA\cite{kim2024openvla} & 11.24 $\pm$ 2.89 & 3.60 $\pm$ 0.55 & 95.00 \\
 & STATIC\cite{haon2025mechanistic} & 11.37 $\pm$ 1.41 & 3.64 $\pm$ 0.57 & 100.00 \\
 & Ours (PID) & 11.46 $\pm$ 1.63 & 3.53 $\pm$ 0.79 & 100.00 \\
 & Ours (RL) & 11.75 $\pm$ 2.65 & 3.49 $\pm$ 0.74 & 100.00 \\
\midrule
\multirow{4}{*}{\makecell[l]{Put the bowl on \\the plate}} & OpenVLA\cite{kim2024openvla} & 18.16 $\pm$ 1.05 & 3.55 $\pm$ 0.51 & 95.00 \\
 & STATIC\cite{haon2025mechanistic} & 17.16 $\pm$ 3.54 & 3.45 $\pm$ 0.37 & 95.00 \\
 & Ours (PID) & 17.92 $\pm$ 0.96 & 3.35 $\pm$ 0.47 & 90.00 \\
 & Ours (RL) & 18.23 $\pm$ 1.04 & 3.69 $\pm$ 0.43 & 90.00 \\
\midrule
\multirow{4}{*}{\makecell[l]{Put the wine bottle \\on the rack}} & OpenVLA\cite{kim2024openvla} & 8.85 $\pm$ 3.70 & 3.60 $\pm$ 0.55 & 65.00 \\
 & STATIC\cite{haon2025mechanistic} & 9.77 $\pm$ 1.45 & 3.69 $\pm$ 0.59 & 70.00 \\
 & Ours (PID) & 9.34 $\pm$ 1.58 & 3.41 $\pm$ 0.78 & 55.00 \\
 & Ours (RL) & 9.53 $\pm$ 2.46 & 3.13 $\pm$ 0.86 & 65.00 \\
\midrule
\bottomrule
\end{tabular}
}
\label{tab:goal_speed}
\end{table*}

\begin{table*}[t]
\centering
\caption{Detailed task-wise results on Speed intervention for LIBERO OBJECT tasksuite}
\small
\scalebox{0.9}{
\begin{tabular}{llccc}
\toprule
Task & Method & Speed (cm/s) & SAT & SR (\%) \\
\midrule
\multirow{4}{*}{\makecell[l]{Pick the alphabet\\ soup and place it \\in the basket}} & OpenVLA\cite{kim2024openvla} & 14.05 $\pm$ 2.54 & 2.16 $\pm$ 0.31 & 90.00 \\
 & STATIC\cite{haon2025mechanistic} & 16.61 $\pm$ 1.48 & 2.23 $\pm$ 0.24 & 85.00 \\
 & Ours (PID) & 15.87 $\pm$ 1.42 & 2.20 $\pm$ 0.29 & 85.00 \\
 & Ours (RL) & 14.55 $\pm$ 2.70 & 2.22 $\pm$ 0.32 & 70.00 \\
\midrule
\multirow{4}{*}{\makecell[l]{Pick up the cream \\cheese and \\ place it \\in the  basket}} & OpenVLA\cite{kim2024openvla} & 13.93 $\pm$ 2.06 & 2.30 $\pm$ 0.31 & 70.00 \\
 & STATIC\cite{haon2025mechanistic} & 13.08 $\pm$ 1.85 & 2.39 $\pm$ 0.26 & 60.00 \\
 & Ours (PID) & 13.11 $\pm$ 3.21 & 2.40 $\pm$ 0.21 & 60.00 \\
 & Ours (RL) & 13.72 $\pm$ 2.83 & 2.48 $\pm$ 0.26 & 75.00 \\
\midrule
\multirow{4}{*}{\makecell[l]{Pick up the salad\\ dressing and \\ place it \\in the basket}} & OpenVLA\cite{kim2024openvla} & 13.87 $\pm$ 2.96 & 1.94 $\pm$ 0.18 & 75.00 \\
 & STATIC\cite{haon2025mechanistic} & 14.68 $\pm$ 1.19 & 2.11 $\pm$ 0.24 & 75.00 \\
 & Ours (PID) & 13.77 $\pm$ 2.04 & 2.00 $\pm$ 0.22 & 85.00 \\
 & Ours (RL) & 14.66 $\pm$ 1.25 & 1.92 $\pm$ 0.34 & 75.00 \\
\midrule
\multirow{4}{*}{\makecell[l]{Pick up the bbq \\sauce and  place it in \\the  basket}} & OpenVLA\cite{kim2024openvla} & 13.02 $\pm$ 1.15 & 1.86 $\pm$ 0.34 & 40.00 \\
 & STATIC\cite{haon2025mechanistic} & 11.37 $\pm$ 3.31 & 1.88 $\pm$ 0.32 & 60.00 \\
 & Ours (PID) & 11.26 $\pm$ 2.92 & 1.75 $\pm$ 0.40 & 40.00 \\
 & Ours (RL) & 11.82 $\pm$ 3.22 & 1.63 $\pm$ 0.35 & 65.00 \\
\midrule
\multirow{4}{*}{\makecell[l]{Pick up the ketchup and \\ place it in the \\ basket}} & OpenVLA\cite{kim2024openvla} & 15.01 $\pm$ 2.20 & 1.92 $\pm$ 0.28 & 90.00 \\
 & STATIC\cite{haon2025mechanistic} & 15.20 $\pm$ 2.30 & 2.10 $\pm$ 0.17 & 65.00 \\
 & Ours (PID) & 14.91 $\pm$ 2.18 & 2.04 $\pm$ 0.35 & 90.00 \\
 & Ours (RL) & 13.62 $\pm$ 3.61 & 2.07 $\pm$ 0.26 & 95.00 \\
\midrule
\multirow{4}{*}{\makecell[l]{Pick up the tomato\\ sauce and  place it \\in the basket}} & OpenVLA\cite{kim2024openvla} & 12.17 $\pm$ 2.32 & 2.07 $\pm$ 0.30 & 75.00 \\
 & STATIC\cite{haon2025mechanistic} & 13.28 $\pm$ 2.75 & 2.11 $\pm$ 0.29 & 100.00 \\
 & Ours (PID) & 11.90 $\pm$ 2.53 & 2.20 $\pm$ 0.48 & 85.00 \\
 & Ours (RL) & 12.08 $\pm$ 2.47 & 1.89 $\pm$ 0.48 & 70.00 \\
\midrule
\multirow{4}{*}{\makecell[l]{Pick up the butter \\and  place it in the \\ basket}} & OpenVLA\cite{kim2024openvla} & 16.24 $\pm$ 1.39 & 2.57 $\pm$ 0.33 & 60.00 \\
 & STATIC\cite{haon2025mechanistic} & 16.52 $\pm$ 1.95 & 3.02 $\pm$ 0.17 & 70.00 \\
 & Ours (PID) & 16.84 $\pm$ 1.34 & 2.82 $\pm$ 0.38 & 55.00 \\
 & Ours (RL) & 16.55 $\pm$ 0.93 & 2.79 $\pm$ 0.20 & 80.00 \\
\midrule
\multirow{4}{*}{\makecell[l]{Pick up the milk and \\ place it in the \\ basket}} & OpenVLA\cite{kim2024openvla} & 15.79 $\pm$ 3.06 & 2.45 $\pm$ 0.31 & 70.00 \\
 & STATIC\cite{haon2025mechanistic} & 16.78 $\pm$ 0.81 & 2.52 $\pm$ 0.28 & 85.00 \\
 & Ours (PID) & 16.72 $\pm$ 1.52 & 2.52 $\pm$ 0.25 & 100.00 \\
 & Ours (RL) & 15.40 $\pm$ 3.28 & 2.46 $\pm$ 0.23 & 95.00 \\
\midrule
\multirow{4}{*}{\makecell[l]{Pick up the pudding and \\ place it in the \\ basket}} & OpenVLA\cite{kim2024openvla} & 16.23 $\pm$ 1.75 & 2.42 $\pm$ 0.17 & 60.00 \\
 & STATIC\cite{haon2025mechanistic} & 15.98 $\pm$ 2.04 & 2.50 $\pm$ 0.19 & 90.00 \\
 & Ours (PID) & 16.60 $\pm$ 1.26 & 2.40 $\pm$ 0.19 & 65.00 \\
 & Ours (RL) & 15.10 $\pm$ 2.06 & 2.49 $\pm$ 0.23 & 50.00 \\
\midrule
\multirow{4}{*}{\makecell[l]{Pick up the orange\\ juice and  place it in\\ the basket}} & OpenVLA\cite{kim2024openvla} & 11.86 $\pm$ 4.75 & 1.73 $\pm$ 0.25 & 90.00 \\
 & STATIC\cite{haon2025mechanistic} & 14.36 $\pm$ 2.57 & 1.80 $\pm$ 0.21 & 95.00 \\
 & Ours (PID) & 14.32 $\pm$ 1.69 & 1.70 $\pm$ 0.28 & 100.00 \\
 & Ours (RL) & 13.31 $\pm$ 3.76 & 1.77 $\pm$ 0.30 & 90.00 \\
\midrule
\bottomrule
\end{tabular}
}

\label{tab:object_speed}
\end{table*}

\begin{table*}[t]
\centering
\caption{Detailed task-wise results on Speed intervention for LIBERO SPATIAL tasksuite}
\small
\scalebox{0.9}{
\begin{tabular}{llccc}
\toprule
Task & Method & Speed (cm/s) & SAT & SR (\%) \\
\midrule
\multirow{4}{*}{\makecell[l]{Pick the akita black bowl \\ between the plate and the \\ramekin and place it on\\ the plate}} & OpenVLA\cite{kim2024openvla} & 17.73 $\pm$ 2.30 & 4.33 $\pm$ 0.33 & 75.00 \\
 & STATIC\cite{haon2025mechanistic} & 17.96 $\pm$ 2.59 & 4.32 $\pm$ 0.32 & 90.00 \\
 & Ours (PID) & 18.60 $\pm$ 1.13 & 4.30 $\pm$ 0.35 & 95.00 \\
 & Ours (RL) & 18.10 $\pm$ 3.69 & 4.44 $\pm$ 0.36 & 80.00 \\
\midrule
\multirow{4}{*}{\makecell[l]{Pick the akita black bowl next\\ to the ramekin and place\\ it on the plate}} & OpenVLA\cite{kim2024openvla} & 18.57 $\pm$ 1.28 & 3.46 $\pm$ 0.61 & 95.00 \\
 & STATIC\cite{haon2025mechanistic} & 18.88 $\pm$ 2.02 & 3.63 $\pm$ 0.40 & 90.00 \\
 & Ours (PID) & 18.03 $\pm$ 3.03 & 3.47 $\pm$ 0.34 & 95.00 \\
 & Ours (RL) & 18.93 $\pm$ 1.27 & 3.44 $\pm$ 0.32 & 90.00 \\
\midrule
\multirow{4}{*}{\makecell[l]{Pick the akita black bowl from\\ table center and place it\\ on the plate}} & OpenVLA\cite{kim2024openvla} & 14.08 $\pm$ 3.45 & 2.55 $\pm$ 0.30 & 95.00 \\
 & STATIC\cite{haon2025mechanistic} & 14.71 $\pm$ 2.30 & 2.45 $\pm$ 0.38 & 65.00 \\
 & Ours (PID) & 14.66 $\pm$ 3.37 & 2.61 $\pm$ 0.33 & 75.00 \\
 & Ours (RL) & 14.55 $\pm$ 3.35 & 2.50 $\pm$ 0.36 & 85.00 \\
\midrule
\multirow{4}{*}{\makecell[l]{Pick the akita black bowl next \\to the cookies box and \\place it on the plate}} & OpenVLA\cite{kim2024openvla} & 19.49 $\pm$ 1.24 & 3.89 $\pm$ 0.34 & 85.00 \\
 & STATIC\cite{haon2025mechanistic} & 18.49 $\pm$ 3.24 & 3.88 $\pm$ 0.27 & 95.00 \\
 & Ours (PID) & 19.22 $\pm$ 1.97 & 4.02 $\pm$ 0.34 & 100.00 \\
 & Ours (RL) & 19.38 $\pm$ 1.37 & 3.94 $\pm$ 0.29 & 100.00 \\
\midrule
\multirow{4}{*}{\makecell[l]{Pick the akita black bowl in \\the top layer of the wooden \\cabinet and place it on\\ the plate}} & OpenVLA\cite{kim2024openvla} & 17.24 $\pm$ 0.55 & 3.39 $\pm$ 0.50 & 70.00 \\
 & STATIC\cite{haon2025mechanistic} & 15.63 $\pm$ 4.66 & 3.22 $\pm$ 0.28 & 75.00 \\
 & Ours (PID) & 16.37 $\pm$ 2.67 & 3.44 $\pm$ 0.46 & 75.00 \\
 & Ours (RL) & 17.32 $\pm$ 0.59 & 3.25 $\pm$ 0.41 & 65.00 \\
\midrule
\multirow{4}{*}{\makecell[l]{Pick the akita black bowl on\\ the ramekin and place it on\\ the plate}} & OpenVLA\cite{kim2024openvla} & 18.29 $\pm$ 2.91 & 3.48 $\pm$ 0.38 & 40.00 \\
 & STATIC\cite{haon2025mechanistic} & 17.59 $\pm$ 2.23 & 3.45 $\pm$ 0.36 & 45.00 \\
 & Ours (PID) & 18.32 $\pm$ 2.09 & 3.29 $\pm$ 0.50 & 50.00 \\
 & Ours (RL) & 17.18 $\pm$ 3.21 & 3.30 $\pm$ 0.53 & 50.00 \\
\midrule
\multirow{4}{*}{\makecell[l]{Pick the akita black bowl on\\ the cookies box and place \\it on the plate}} & OpenVLA\cite{kim2024openvla} & 17.87 $\pm$ 1.17 & 3.30 $\pm$ 0.35 & 95.00 \\
 & STATIC\cite{haon2025mechanistic} & 18.19 $\pm$ 0.85 & 3.30 $\pm$ 0.28 & 95.00 \\
 & Ours (PID) & 17.99 $\pm$ 1.12 & 3.30 $\pm$ 0.26 & 85.00 \\
 & Ours (RL) & 18.04 $\pm$ 1.49 & 3.31 $\pm$ 0.34 & 100.00 \\
\midrule
\multirow{4}{*}{\makecell[l]{Pick the akita black bowl\\ on the stove and place it\\ on the plate}} & OpenVLA\cite{kim2024openvla} & 15.35 $\pm$ 2.98 & 3.44 $\pm$ 0.60 & 70.00 \\
 & STATIC\cite{haon2025mechanistic} & 15.55 $\pm$ 0.93 & 2.90 $\pm$ 0.89 & 65.00 \\
 & Ours (PID) & 16.01 $\pm$ 0.82 & 2.92 $\pm$ 0.90 & 70.00 \\
 & Ours (RL) & 15.54 $\pm$ 1.98 & 3.28 $\pm$ 0.82 & 65.00 \\
\midrule
\multirow{4}{*}{\makecell[l]{Pick the akita black bowl\\ next to the plate and\\ place it on the plate}} & OpenVLA\cite{kim2024openvla} & 17.64 $\pm$ 1.55 & 3.37 $\pm$ 0.68 & 90.00 \\
 & STATIC\cite{haon2025mechanistic} & 16.30 $\pm$ 3.30 & 3.48 $\pm$ 0.58 & 70.00 \\
 & Ours (PID) & 16.75 $\pm$ 2.48 & 3.37 $\pm$ 0.60 & 75.00 \\
 & Ours (RL) & 17.49 $\pm$ 2.60 & 3.48 $\pm$ 0.74 & 75.00 \\
\midrule
\multirow{4}{*}{\makecell[l]{Pick the akita black bowl on \\the wooden cabinet and place \\it on the plate}} & OpenVLA\cite{kim2024openvla} & 19.67 $\pm$ 2.11 & 4.97 $\pm$ 0.33 & 65.00 \\
 & STATIC\cite{haon2025mechanistic} & 19.81 $\pm$ 1.11 & 4.94 $\pm$ 0.35 & 75.00 \\
 & Ours (PID) & 20.08 $\pm$ 0.61 & 4.93 $\pm$ 0.23 & 60.00 \\
 & Ours (RL) & 18.83 $\pm$ 3.16 & 5.04 $\pm$ 0.36 & 75.00 \\
\midrule
\bottomrule
\end{tabular}
}

\label{tab:spatial_speed}
\end{table*}

\end{document}